\documentclass{article} 
\usepackage{iclr2026_conference,times}

\usepackage{amsmath,amsfonts,bm}

\def\eqref#1{equation~\ref{#1}}

\def\1{\bm{1}}

\DeclareMathAlphabet{\mathsfit}{\encodingdefault}{\sfdefault}{m}{sl}
\SetMathAlphabet{\mathsfit}{bold}{\encodingdefault}{\sfdefault}{bx}{n}

\usepackage{hyperref}
\usepackage{url}
\usepackage{booktabs}
\usepackage{xspace}
\usepackage{subcaption}
\usepackage{multirow}
\usepackage{colortbl}
\usepackage{booktabs}
\usepackage{multirow}
\usepackage{adjustbox} 
\usepackage[table]{xcolor}
\usepackage{rotating}
\usepackage{graphicx}
\usepackage{threeparttable}
\usepackage[textsize=tiny]{todonotes}
\usepackage{graphicx}
\usepackage{algorithm}
\usepackage{algpseudocode}
\usepackage{listings}
\usepackage{listings-ext}
\usepackage{wrapfig}
\usepackage{titletoc}
\usepackage{enumitem}
\usepackage{wrapfig}

\title{SCRIBES: Web-Scale Script-Based Semi-Structured Data Extraction with Reinforcement Learning}

\author{Shicheng Liu$^{1}$\thanks{Work done at Meta}, Kai Sun$^{2}$, Lisheng Fu$^{2}$, Xilun Chen$^{3}$, Xinyuan Zhang$^{2}$, Zhaojiang Lin$^{2}$, \\ 
\textbf{Rulin Shao$^{3,4}$, Yue Liu$^{2}$}\textbf{, Anuj Kumar$^{2}$, Wen-tau Yih$^{3}$, Xin Luna Dong$^{2}$}\\
$^1$ Stanford University,\quad ~~~$^2$ Meta Reality Labs,\\
$^3$ FAIR at Meta, \quad $^4$ University of Washington \\
\texttt{shicheng@cs.stanford.edu, sunkaicn@meta.com}
}

\newcommand{\system}{\textsc{SCRIBES}\xspace}

\definecolor{veryhigh}{RGB}{212, 237, 218}    
\definecolor{high}{RGB}{231, 243, 231}        
\definecolor{neutral}{RGB}{255, 243, 205}     
\definecolor{low}{RGB}{248, 215, 218}         
\definecolor{verylow}{RGB}{245, 198, 203}     
\definecolor{llamashade}{RGB}{220,240,255}
\definecolor{gptshade}{RGB}{255,230,230}
\definecolor{mediumgreen}{RGB}{60, 179, 113}
\definecolor{darkgreen}{rgb}{0.0, 0.5, 0.0}

\lstdefinelanguage{Jinja2}{
  morekeywords={},
  sensitive=false,
  moredelim=[s][\color{blue}]{\{}{\}},
  moredelim=[s][\color{blue}]{\%}{\%},
  moredelim=[s][\color{mediumgreen}]{\{####}{####\}},
}
\lstdefinestyle{python}{
    language=Python,
    basicstyle=\ttfamily\small,
    backgroundcolor=\color{gray!5},
    keywordstyle=\color{blue}\bfseries,
    commentstyle=\color{gray}\itshape,
    stringstyle=\color{orange!70!black},
    showstringspaces=false,
    frame=single,
    rulecolor=\color{black!20},
    tabsize=4,
    breaklines=true,
    breakatwhitespace=true,
}

\iclrfinalcopy
\begin{document}

\maketitle

\begin{abstract}


Semi-structured content in HTML tables, lists, and infoboxes accounts for a substantial share of factual data on the web, yet the formatting complicates usage, and reliably extracting structured information from them remains challenging. Existing methods either lack generalization or are resource-intensive due to per-page LLM inference. In this paper, we introduce \system (\textbf{SCRI}pt-\textbf{B}ased Semi-Structured Content \textbf{E}xtraction at Web-\textbf{S}cale), a novel reinforcement learning framework that leverages layout similarity across webpages within the same site as a reward signal. Instead of processing each page individually, \system generates reusable extraction scripts that can be applied to groups of structurally similar webpages. Our approach further improves by iteratively training on synthetic annotations from in-the-wild CommonCrawl data. Experiments show that our approach outperforms strong baselines by over 13\% in script quality and boosts downstream question answering accuracy by more than 4\% for GPT-4o, enabling scalable and resource-efficient web information extraction.\footnote{Code will be released at \url{https://github.com/facebookresearch/SCRIBES}.}

\end{abstract}


\section{Introduction}

A substantial volume of web data is stored in semi-structured formats such as HTML (HyperText Markup Language) tables, lists, and infoboxes~\citep{10.1145/2623330.2623623, kai-2025-arxiv}\footnote{See Appendix~\ref{sec:semi-structured_content_appendix} for a discussion of different types of webpages with semi-structured content.}. Such content offers a rich source of factual information, yet its formatting complicates effective usage in downstream applications like question answering~\citep{tan2024htmlrag, kai-2025-arxiv}. Knowledge extraction aims to transform such data from raw HTML into structured representations (e.g., triples)~\citep{10.5555/645856.758279}, but despite decades of research, this remains a major challenge at large scale.
Existing approaches fall into two main categories. {\em Traditional information extraction (IE) methods}, such as wrapper induction~\citep{kushmerick1997wrapper}, graph mining~\citep{10.5555/645927.672370, 10.1145/956750.956826}, layout-based methods~\citep{10.1145/1060745.1060761, 10.14778/3231751.3231758}, and Deep Neural Networks~\citep{dalvi2011automatic, lockard-etal-2020-zeroshotceres}, tend to be brittle and struggle to generalize over unseen data or schema.
More recently, Large Language Model (LLM)-based methods have emerged that parse individual pages or construct Knowledge Graphs (KGs) using large models~\citep{gutiérrez2024hipporag, zhang-soh-2024-extract, ning2023uukg, 10386454, zhang-etal-2023-aligning, bai2025autoschemakgautonomousknowledgegraph}. Although these methods can produce high-quality outputs, they are resource-intensive to apply at scale because they require invoking an LLM for every page.

{\em Can we extract knowledge from semi-structured content at the web scale both effectively and efficiently?} In this paper, we 
introduce {\bf \system: \underline{SCRI}pt-\underline{B}ased Semi-Structured Content \underline{E}xtraction at Web-\underline{S}cale}, a novel approach for large-scale knowledge extraction. Given a webpage, \system\ leverages an LLM to generate an extraction script that applies to other pages within the same domain, which typically share highly similar layouts (Figure~\ref{fig2:similar_pages_demo}). Executing the script incurs only negligible resource cost compared with running an LLM-based extraction on every individual page.

Although the idea appears straightforward, current LLMs struggle to produce high-quality, generalizable extraction scripts. Fine-tuning them for this ability is cumbersome, as creating annotations for such scripts is difficult even for expert labelers. The success of \system\ lies in a Reinforcement Learning (RL) framework that leverages structural similarities across related webpages: given a group of similar webpages, the model is rewarded when a script generated for one webpage also works on others. This encourages learning scripts that generalize beyond individual examples.

\begin{figure}[!t]
    \centering
    \includegraphics[width=0.83\textwidth]{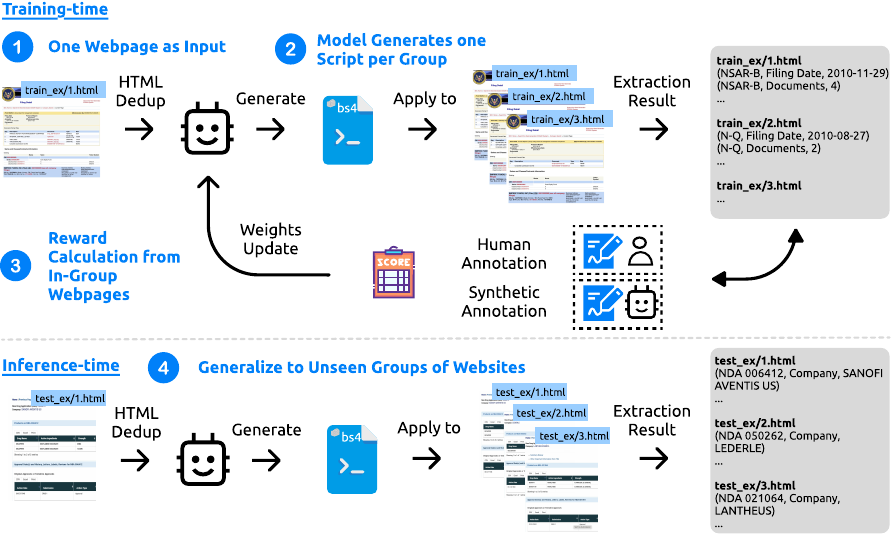}
    \caption{\system organizes similar webpages into groups under each website. During training, the model receives one representative webpage per group as input (pt. 1) and is tasked with generating a single extraction script applicable to all similar webpages within the group (pt. 2). Extraction results are then compared against human annotations for labeled data and synthetic annotations for unlabeled CommonCrawl webpages. The resulting scores are used to update the model weights (pt. 3). At inference time, \system enables the model to generalize to \textit{new, unseen websites} by generating scripts that can be applied across similar webpages (pt. 4).}
    \label{fig1:overview}
\end{figure}

\system\ draws training data from two sources. First, it learns from {\em a small set of annotated examples} (192 pages from 34 groups) (Figure~\ref{fig1:overview}, parts 1–3). For each group, \system\ takes one webpage as input and prompts the model to generate a script intended to generalize across the group. The script is then executed on the remaining pages, and its outputs are compared with annotations to compute the reward. Second, \system\ leverages {\em in-the-wild websites from CommonCrawl} to further enhance its capabilities. We develop an iterative approach that starts from a checkpoint trained on annotated data and then refines the model to continue learning from their failed predictions on the in-the-wild websites. To provide supervision at scale, we employ LLM-based direct extractions as synthetic annotations, reducing reliance on annotations or hand-crafted parsers.

Extensive experiments show that our RL-trained model outperforms strong agentic baselines by more than 13\% in generating robust, reusable parsing scripts.  Moreover, we demonstrate that {\em improved extraction translates into downstream benefits}: in QA tasks requiring structured reasoning over HTML, incorporating triples produced by \system\ boosts accuracy across a wide range of LLMs, including SOTA models such as GPT-4o by over 4\%. 
\section{Related Works}


\subsection{Semi-Structured Data Processing}

\textbf{Flattening}: In complex QA or retrieval settings that mix texts, tables, and knowledge bases, a common practice is to ``linearize'' everything into plain text~\citep{oguz-etal-2022-unik, zhang-etal-2024-spaghetti, ma-etal-2022-open, 10.1145/3477495.3531815}. This is also a popular practice when dealing with HTML pages. Trafilatura is a widely used HTML cleaning and text extraction toolkit designed for large-scale web processing~\citep{barbaresi-2021-trafilatura}, among many other HTML conversion packages~\citep{firecrawl, newspaper4k}. While effective for general text extraction, these utilities typically discard or flatten structural elements such as tables, lists, and infoboxes. Similar to findings in complex QA that highlight the importance of structural cues~\citep{liu-etal-2024-suql, zhang-etal-2024-spaghetti}, recent work on RAG with raw HTML shows that converting to plain text discards headings, table structures, and other layout information critical for downstream tasks~\citep{tan2024htmlrag}. 

\textbf{Traditional IE Methods}: A classical approach to extracting structured data from semi-structured web content is wrapper induction, which learns extraction procedures (“wrappers”) from a small set of labeled examples instead of hand-crafted rules \citep{kushmerick1997wrapper}. Extensions include boosted wrapper induction, which combines simple patterns for greater robustness \citep{freitag2000boosted}, and large-scale methods that handle noisy data and template drift \citep{dalvi2011automatic}. While effective on regular site structures with clean annotations, these methods are brittle to structural changes and generalize poorly across diverse domains. {In contrast, our approach learns executable scripts, i.e. full extraction programs that operate directly on raw HTML, allowing the system to generalize beyond fixed rules and adapt automatically without manual template design.}



\textbf{LLM-based methods:} Several recent advances utilize LLMs to extract semi-structured contents. For instance, \citet{wang2025readerlmv2smalllanguagemodel} train a LLM to convert HTMLs into Markdown and JSON using SFT and RL methods. Similarly, \citet{olmocr} use a VLM to convert PDFs into clean, readable format retaining tabular structures. Many related works also exist on LLM-assisted knowledge-base construction~\citep{gutiérrez2024hipporag, zhang-soh-2024-extract, ning2023uukg, 10386454,zhang-etal-2023-aligning,bai2025autoschemakgautonomousknowledgegraph}. However, calling an LLM per page remains resource-intensive at web-scale; moreover, they typically treat each page independently, missing the cross-page layout regularities that \system exploits.

\subsection{RL Without Annotations}

A growing body of work explores reinforcement learning in settings without explicit annotations. \citet{zuo2025ttrl} show that models can refine themselves at test time by turning consensus among rollouts into rewards, while \citet{zhao2025learning} and \citet{prabhudesai2025maximizing} demonstrate that internal signals such as self-certainty or confidence are sufficient to drive continued improvement. \citet{shao2025spurious} find that even spurious or random rewards can produce surprising gains, suggesting that models can bootstrap from imperfect signals. Like prior work, we reduce dependence on annotations by iteratively refining the model from its own failures, but instead of relying solely on internal signals, we utilize LLM-based direct extractions as synthetic annotation for reward calculation.

\section{\system Framework}
\subsection{Problem Definition}

\begin{figure}[ht]

    \centering
    \includegraphics[width=0.9\textwidth]{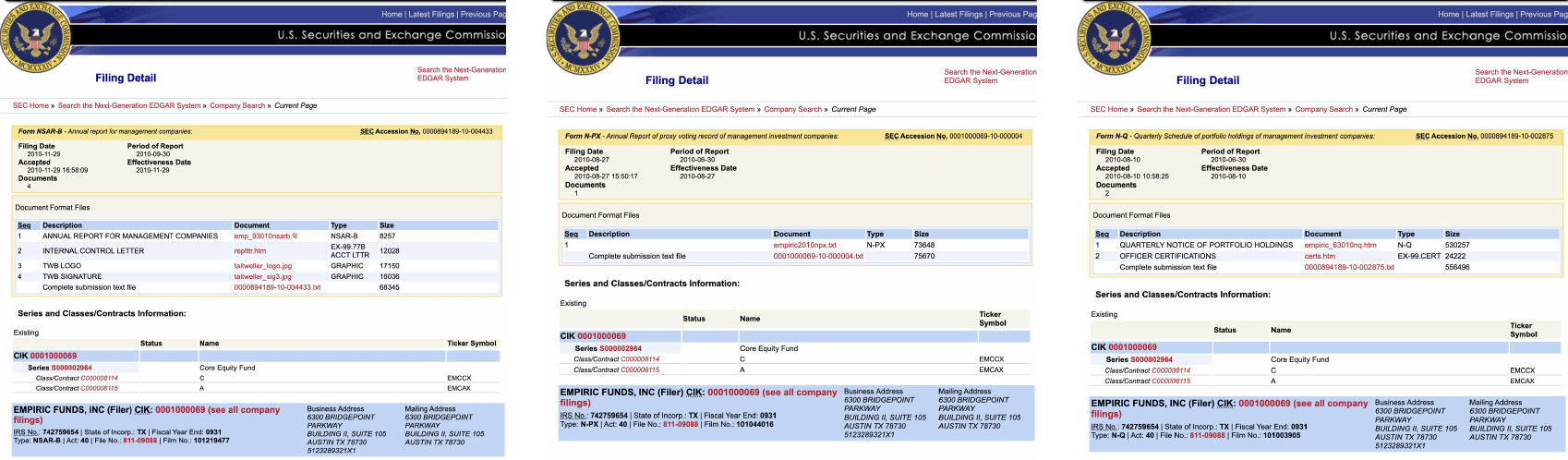}
    \caption{Three webpages containing semi-structured content under the same website.}
    \label{fig2:similar_pages_demo}
    \vspace{-10pt}
\end{figure}

\paragraph{Knowledge extraction:} Let $G=\{p_1, \cdots, p_n\}$ be a group of semi-structured webpages that are structurally similar. The {\em knowledge extraction} task parses each page $p_i, i \in [1,n],$ to a list of triples (subjects, predicates, and objects). We denote by $y^{\star}_{p_i}$ the ground truth triples for page $p_i$.


\paragraph{Extraction script generation:} We propose to solve the knowledge extraction problem by generating an extraction script that applies to every page in $G$. Formally, our goal is to train a model $LM$ that, given any webpage $p \in G$, predicts an extraction script $\hat{y}_p = LM(p)$, such that applying $\hat{y}$ to every page in $G$ generates triples close to ground truth triples $\{y^{\star}_{p_i} | p_i \in G\}$. For instance, in \autoref{fig2:similar_pages_demo}, a model-generated script should robustly handle variations across webpages, such as differences in table sizes and values.


\subsection{HTML Deduplication (Dedup)}

The raw HTMLs of webpages are typically very long and can easily surpass the maximum context window of even the long-context LLMs. We propose a simple yet effective method for deduplicating HTMLs: repeated HTML blocks are collapsed into a compact representation of the form ``$n$ more \dots elements,’’ which substantially reduces context length. Ablation experiments confirm that this deduplication step significantly improves model performance. We therefore apply it throughout our \system-trained models. An example of the dedup process is shown in Figure \ref{fig:dedup_example}, and further details and analysis are provided in Appendix~\ref{sec:appendix-dedup-algo}.

\subsection{RL Setup}
\label{sec:rl_setup}

Annotating such extraction scripts for training is challenging even for expert human annotators. To address this, rather than relying on demonstrations, we propose adopting \textit{Reinforcement Learning with Verifiable Rewards (RLVR)} for this task.

We define $r(p\!\to\!q) = S\!\bigl(\hat{y}_{p}(q),\,y^{\star}_q\bigr)\in[0,1]$ as the score obtained when the script $\hat{y}_p$ is executed on a (possibly different) page $q$, where $S$ is a scoring function that measures similarity between predicted and annotated tuples. To compute this score, we follow prior works \citep{liu-etal-2024-spinach, kai-2025-arxiv} and adopt a bipartite matching algorithm that aligns predicted triples with gold triples by maximizing their pairwise fuzzy matching score. Based on this matching, we compute fuzzy precision $P^{\mathrm{fuzzy}}$, recall $R^{\mathrm{fuzzy}}$, and $F_1$ score $F_1^{\mathrm{fuzzy}}$. Since fuzzy string similarity may fail to fully capture semantic equivalence, we additionally employ an LLM-as-a-judge (set to \texttt{Llama-3.3-70B-Instruct}) to evaluate the aligned triples (Prompt \ref{prompt:llm-as-a-judge}). We choose Llama to ensure consistency with prior work~\citep{kai-2025-arxiv} and, by fixing the checkpoint, to enable reproducible experiments. This yields LLM-based precision $P^{\mathrm{LM}}$, recall $R^{\mathrm{LM}}$, and $F_1$ score $F_1^{\mathrm{LM}}$. During training, we set $S = F_1^{\text{fuzzy}}$, the triple-level fuzzy $F_1$ score. Refer to Appendix \ref{sub:appendix_metrics} for additional details on metrics and an optimized implementation of $F_1^{\text{fuzzy}}$ during training.  

\subsubsection{Reward Signal from Labeled Data}
\label{sec:reward_signal_labeled}

We define the following notations:

\begin{enumerate}
    \item the \textit{self-score} is $r_{\text{self}}(p)=r(p\!\to\!p)$, while
    \item each \textit{cross-score} is $r_{\text{cross}}(p,q)=r(p\!\to\!q)$ for $q\neq p$.
\end{enumerate}

\system optimizes a model using Group Relative Policy Optimization (GRPO)~\citep{shao2024deepseekmathpushinglimitsmathematical} based on the following reward function for each training sample $p$:
\begin{equation}
\label{eq_scribes}
r_{\system}(p)=\tfrac{1}{|G(p)|}\sum_{q\in G(p)}r(p\!\to\!q) = \tfrac{1}{|G(p)|} r_{\text{self}}(p) +  \tfrac{|G(p)| - 1}{|G(p)|}\sum_{q \in G(p), p \ne q} r_{\text{cross}}(p,q)
\end{equation}
Within this framework, each self-score contributes only $\tfrac{1}{|G(p)|}$ to the final reward, while cross-scores constitute the majority of the reward signal. This design strongly encourages the model to generalize by accounting for potential variations across other, unseen webpages within the same group. We study the effect of different reward formulations through ablation studies in Section~\ref{sec:ablations}.

\subsubsection{Reward Signal from Unlabeled Data in the Wild}
\label{sec:reward_signal_unlabeled}

\begin{figure}[!t]
    \centering
    \includegraphics[width=\textwidth]{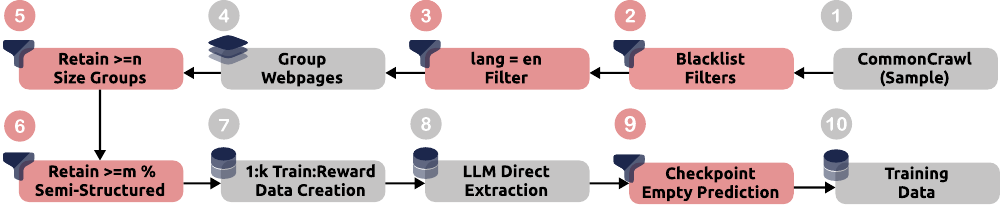}
    \caption{Processing pipeline for unlabeled data from CommonCrawl in Section \ref{sec:reward_signal_unlabeled}.}
    \label{fig:commoncrawl}
    \vspace{-10pt}
\end{figure}

When training on annotated data, \system can directly leverage the gold human annotation \(y_p^\star\) for each page \(p\) as the reward signal. However, because the only high-quality annotated dataset available from \citet{kai-2025-arxiv} is relatively small, it is inherently difficult to achieve broad coverage of diverse website layouts using annotated data alone. To address this limitation, we propose a novel approach that leverages unlabeled in-the-wild webpages from CommonCrawl (abbreviated as CC)~\citep{commoncrawl}.

Our data collection pipeline is illustrated in Figure~\ref{fig:commoncrawl}.
\textbf{(pt. 1)} Starting from a sample of CC, \textbf{(pt. 2)} we first apply the blacklist filters from \citet{penedo2024finewebdatasetsdecantingweb} to remove adult or explicit content. \textbf{(pt. 3)} We then apply language filters to select English content websites and \textbf{(pt. 4)} group webpages by domain, \textbf{(pt. 5)} retaining only groups containing at least \(n\) webpages. \textbf{(pt. 6)} Next, we use an LLM-based classifier (Prompt~\ref{prompt:classifier}) to identify webpages containing semi-structured content, and we retain only those website groups where at least \(m\%\) of the pages are classified as semi-structured. \textbf{(pt. 7)} Finally, we sample one webpage as the training example and associate it with up to \(k \leq n\) in-group webpages for reward calculation. In our experiments, we apply the following thresholds: $n=30$, $m=90$, and $k = 13$.

At this stage, we obtain a collection of in-the-wild webpage groups containing semi-structured content. However, without human annotations, it is unclear what reward signal should be used for training. \textbf{(pt. 8)} To address this, we propose using LLM-based direct extraction (Prompt~\ref{prompt:direct-gen}) as a proxy for gold annotations. Our experiments show this to be the strongest baseline. Nevertheless, because such direct extraction is far from perfect (achieving only about 40\% $F_1$ for the best baseline), we aim to prevent noisy rewards from degrading model performance. \textbf{(pt. 9)} To this end, we start from a checkpoint trained on annotated data and identify a subset of webpages where the model's predicted scripts fail to produce any results. By concentrating training on these failure cases, we increase the likelihood that the additional synthetic data improves the model's performance. Ablation studies on the necessity of this subset are presented in Section~\ref{sec:ablations}.







\section{Experiments}

\begin{table}[t]
\centering
\small
\resizebox{\textwidth}{!}{
\begin{tabular}{lccc|ccc|ccc}
\toprule
Model and Method & \multicolumn{3}{c}{All} & \multicolumn{3}{c}{Example}& \multicolumn{3}{c}{Holdout} \\
 \cmidrule(lr){2-4}\cmidrule(lr){5-7}\cmidrule(lr){8-10}
 & $R^{\mathrm{LM}}$ & $P^{\mathrm{LM}}$ & $F_1^{\mathrm{LM}}$ & $R^{\mathrm{LM}}$ & $P^{\mathrm{LM}}$ & $F_1^{\mathrm{LM}}$ & $R^{\mathrm{LM}}$ & $P^{\mathrm{LM}}$ & $F_1^{\mathrm{LM}}$ \\
\midrule
\multicolumn{10}{c}{\cellcolor{gray!10}Baselines (Direct LLM Extraction)} \\

{L-70B~\citep{kai-2025-arxiv} $^*$} & {24.3} & {15.7} & {19.1} & {-} & {-} & {-} & {-} & {-} & {-} \\
{Fine-tuned L-70B~\citep{kai-2025-arxiv} $^*$} & {21.4} & {27.1} & {23.9} & {-} & {-} & {-} & {-} & {-} & {-} \\
{GPT-4o~\citep{kai-2025-arxiv} $^*$} & {35.1} & {23.8} & {28.3} & {-} & {-} & {-} & {-} & {-} & {-} \\

Q-14B flatten & 30.5 & 36.5 & 29.9 & - & -  & -  & -  & -  & -\\
Q-32B flatten & 28.7 & 37.4 & 29.9 & - & -  & -  & -  & -  & - \\
GO-20B 2-shot flatten & 33.2 & \textbf{47.1} & 34.9  & - & -  & -  & -  & -  & -\\
GO-120B 2-shot flatten & \textbf{42.3} & 46.3 & \textbf{40.4}  & - & -  & -  & -  & -  & -\\

\midrule

\multicolumn{10}{c}{\cellcolor{gray!10}Baselines (Script-gen)} \\
Q-14B  agentic-3-iter 2-shot &8.6 & 11.1 & 8.0  & 13.2 & 18.0 & 12.6 & 6.3 & 7.8 & 5.7 \\
L-70B  agentic-3-iter & 10.1 & 15.5 & 10.5 & 16.7 & 23.8 & 16.8 & 6.9 & 11.2 & 7.4 \\
Q-72B  agentic-3-iter 2-shot  & 16.4 & 19.4 & 15.0 & 24.1 & 28.6 & 21.8 & 13.3 & 15.8 & 12.4 \\
Q-32B  agentic-3-iter 2-shot & 18.6 & 27.2 & 19.4  & 24.5 & 34.8 & 25.9 & 15.8 & 23.9  & 16.4 \\
GO-20B  agentic-3-iter  & 24.7 & 23.2 & 20.9 & 29.3 & 26.4 & 27.7 & 22.5 & 21.8  & 18.9 \\
GPT-4o  agentic-3-iter 2-shot & 26.0 & 33.0 & 24.4  & 33.0 & 36.5 & 31.2 & 22.5 & 31.3  & 21.1 \\
GO-120B  agentic-3-iter 2-shot  & \textbf{33.9} & \textbf{41.0} & \textbf{34.3} & \textbf{35.8} & \textbf{42.3} & \textbf{36.6} & \textbf{33.0} & \textbf{40.5} & \textbf{33.3} \\
\midrule

\multicolumn{10}{c}{\cellcolor{blue!10}\system\ (Script-gen)} \\

Q-14B & 23.0 & 24.3 & 19.9  & 31.2 & 29.8 & 26.7 & 19.0 & 21.7 & 16.7 \\
Q-14B (+ CC) & 25.2 & 23.0 & 21.8 &  34.9 & 31.0 & 30.0 & 20.5 & 19.1 & 17.7 \\

Q-32B & 29.9 & 31.5 & 28.1 & 32.0 & 33.9 & 30.3 & 28.8 & 30.3  & 26.8 \\
Q-32B (+ CC) & \textbf{37.4} & \textbf{36.0} &\textbf{33.2} & \textbf{39.5}& \textbf{35.5} & \textbf{34.6} & \textbf{36.2} & \textbf{36.2}  &\textbf{32.4} \\

\bottomrule
\end{tabular}}
\caption{LLM-judged metrics are reported separately for \textit{All}, \textit{Examples} (the webpage model used to generate the script), and \textit{Holdout} (similar webpages where the same script was applied). Columns show macro-averaged $P^{\mathrm{LM}}$, $R^{\mathrm{LM}}$, and $F_1^{\mathrm{LM}}$. For each model and block, we report only the strongest baseline here. The full baseline results, {including LLM-based agentic baselines HippoRAG~\citep{gutiérrez2024hipporag} and AutoSchemaKG~\citep{bai2025autoschemakgautonomousknowledgegraph}, which exhibit lower scores, } are provided in Table~\ref{tab:llm_metrics_all_baselines} in Appendix~\ref{sub:appendiex_full_results}. {($^*$) Numbers reported by \citet{kai-2025-arxiv} are on the full set.}}
\vspace{-10pt}
\label{tab:llm_metrics_example_all}
\end{table}

\subsection{Dataset}

\textbf{Annotated dataset}: Existing datasets for semi-structured knowledge extraction from raw webpages are limited. \textit{SemiBench}~\citep{kai-2025-arxiv} presents a dataset of webpages drawn from 139 popular websites in CommonCrawl, annotated with triples. Their collection includes 83 websites with a single webpage, 46 groups of 3 similar webpages, and 10 groups of 13 similar webpages each. This grouping scheme provides a valuable opportunity to evaluate generalization in the \system setting. We select the 56 groups containing more than 1 webpage each for experiments in this work. We divided the annotated dataset into training and test sets using a 60\%-40\% split \textbf{across groups}; that is, we assign entire groups to either the training or test set, and we do not split within any group. For a group of size $n$ in the training/test set, we create $n$ training/test examples, each using one webpage as input and all group elements used for reward calculation. All evaluation metrics are reported on the test set, which contains only websites from groups that the model did not see during training. Refer to additional details in Appendix \ref{sec:appendix-data-proprocessing}.

\textbf{In-the-wild webpages}: To construct groups directly from CommonCrawl, we employ a simple heuristic: two webpages are grouped together if they share the same URL prefix up to the final substring. For example, \texttt{example.com/mid1/sub1} and \texttt{example.com/mid1/sub2} belong to the same group, while \texttt{example.com/mid2} does not. The LLM used in our pipeline is \texttt{GPT-OSS-120B}. We randomly sampled $50$ webpages and estimated classifier accuracy at $90.0\%$ precision and $72.0\%$ recall. In total, 19,566 groups satisfied the $n \geq 30$ condition, among which 2,003 also satisfied the $m \geq 90$ condition. After direct extraction with the LLM, 1,898 examples were retained (the remainder corresponding to prediction failures or empty outputs). This entire process used less than 1\% of the CC-MAIN-2025-30 crawl. We hypothesize that this pipeline can be scaled to larger portions of CommonCrawl for broader coverage; in this paper, we focus on establishing its feasibility.

\subsection{Training Setup and Baselines}

\textbf{Training} We train \texttt{Qwen2.5-Instruct} family models and perform minimal hyperparameter tuning to ensure stability during model training. Refer to Appendix \ref{sec:appendix-training-setup} for additional details.


\textbf{Baselines} We experiment with both SOTA close-source and open-source models, including: \texttt{gpt-4o}, \texttt{Llama-3.3-70B-instruct} (abbreviated as L-70B), \texttt{Qwen2.5-Instruct} (abbreviated as Q-xB) family, and \texttt{gpt-oss} (abbreviated as GO-xB) family. We implement the following baselines for comparison (Prompt \ref{prompt:script-gen}). By default, all baselines use Dedup as the \system-trained models. We explore multiple configurations to construct strong baseline models.
\begin{enumerate}
    \item \textit{agentic-$n$-iter}: After the model outputs a script given an example, if the script fails to produce output or produces empty output, we feed the execution feedback to the model and ask it to retry. Otherwise we use the output script as prediction. We repeat this ReAct-style~\citep{yao2022react} procedure up to $n$ times;
    \item \textit{$n$-shot}: We feed in $n$ HTMLs and their corresponding gold extraction results as in-context learning examples;
    \item \textit{flatten}: We directly flatten the HTML\footnote{\texttt{BeautifulSoup(html\_content, "html.parser").get\_text()}} and use it as model's input. Note that there is no generalizability requirement or dedup involved in this setup.
    \item {Recent, SOTA LLM-based KG construction pipelines, including HippoRAG~\citep{gutiérrez2024hipporag} and AutoSchemaKG~\citep{bai2025autoschemakgautonomousknowledgegraph}. See Section \ref{sec:appendix-additiona-llm-agent-baselines} for details.}
\end{enumerate}


\subsection{Results}

\textbf{RQ1}: Does \system framework bring improvements to models in terms of their capability to extract semi-structured data?

For each example $p$ in our test set, models generate a script $\hat{y}_p = LM(p)$ and we apply it to all examples in $G(p)$. We derive a score
\begin{equation}
\label{eq:figure1-eval}
S(p) =  \frac{1}{|G(p)|} \sum_{q \in G(p)} S(\hat{y}_p, y^\star_q)
\end{equation}
where we set $S$ to be recall, precision, or $F_1$ score, as defined in Section~\ref{sec:rl_setup}. We refer to this aggregate score as ``All.'' To further investigate the performance gap between the example provided to the model (``Example'') and the other webpages to which the model-generated script is applied (``Holdout''), we decompose the score in Eq.~\ref{eq:figure1-eval} into two separate components:
\begin{equation*}
S_{\texttt{example}}(p) = S(\hat{y}_p, y^\star_p) 
\quad\quad
S_{\texttt{holdout}}(p) = \frac{1}{|G(p)| - 1} \sum_{q \in G(p),\, q \ne p} S(\hat{y}_q, y^\star_q)
\end{equation*}
In Table~\ref{tab:llm_metrics_example_all}, we report the macro average of $R^{\mathrm{LM}}, P^{\mathrm{LM}}, F_1^{\mathrm{LM}}$ by averaging individual $S(p)$ scores. \system-trained models drastically outperform strong agentic baselines. The best Q-14B and Q-32B models outperform the few-shot agentic base model performance by 13.8\% in $F_1^{\mathrm{LM}}$, and our best Q-32B model performs on-par with the few-shot agentic GO-120B model.

\textbf{RQ2}: Does using \system enable resource-efficient, web-scale extraction?

\begin{wrapfigure}{r}{0.5\textwidth}
  \centering
  \vspace{-10pt} 
    \includegraphics[width=0.5\textwidth]{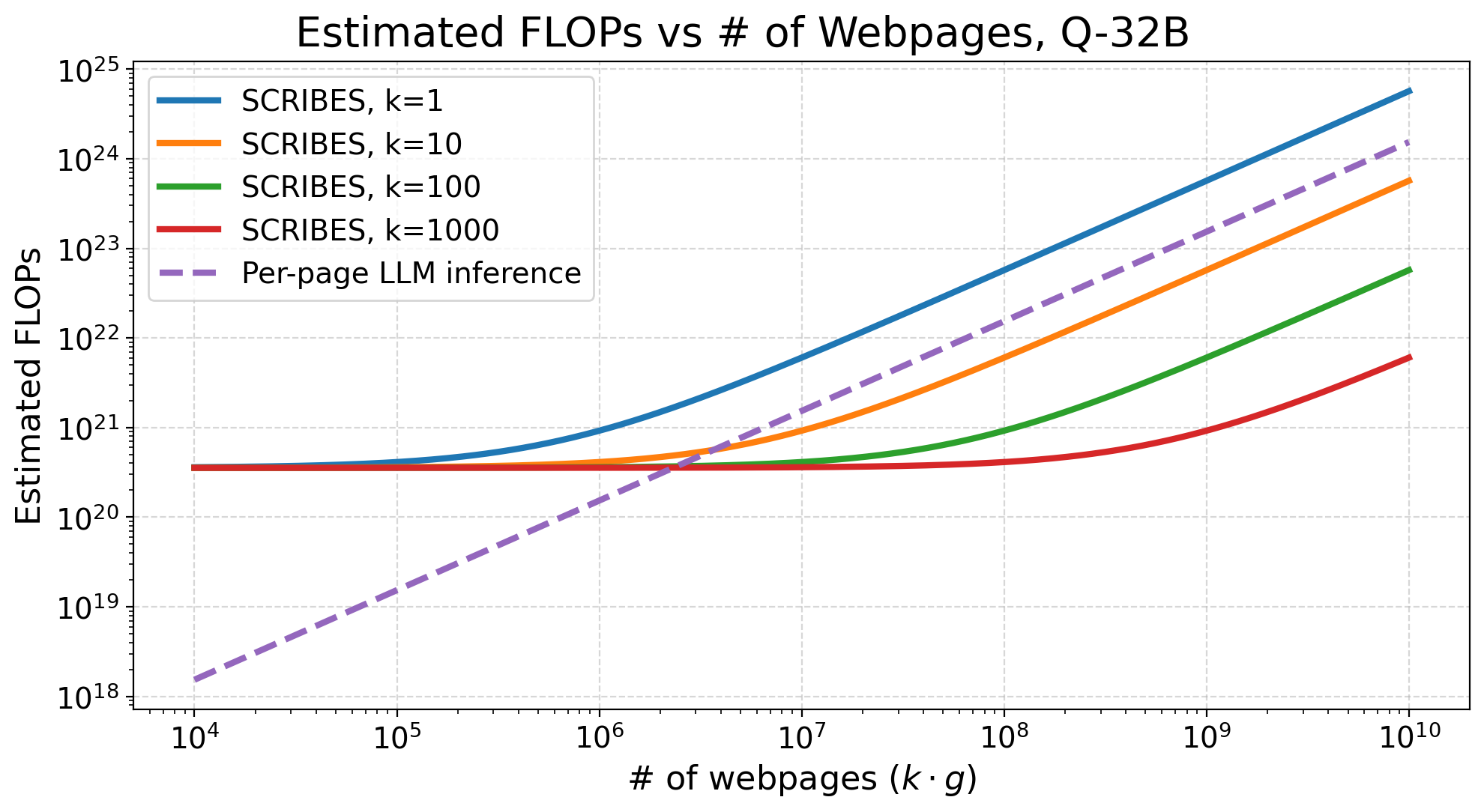}
  \vspace{-10pt} 
    \caption{{Estimated GPU FLOPs usage comparing \system-trained model with per-page LLM inference for Q-32B. The results show that, even when training compute is included, \system-trained models scale more efficiently at web scale.}}
    \label{fig:compute_comparison}
\end{wrapfigure}

To demonstrate the \system-framework's applicability to web-scale semi-structured content extraction, we evaluate on a leftover subset of CommonCrawl data that was not used in model training. To keep the experiment tractable, we capped each group at 30 webpages and required at least 13 webpages per group, meaning this evaluation covers only a tiny fraction of the available data. On this small subset with 113,129 webpages, our model extracted 2,788,760 triples. Remarkably, only 4,661 required direct model predictions, while the vast majority were generated automatically through model-produced scripts.

On average, processing a webpage with deduplicated HTML requires 8,879 tokens, whereas using flattened HTML requires 2,399 tokens. Let $\rho = \frac{8879}{2399} \approx 3.7$ denote this relative per-page token ratio. Our approach quickly becomes more efficient as long as the target website contains at least 4 structurally similar pages. In fact, the token speedup of our scribe-based method relative to flattening grows linearly with $k$ (the number of structurally similar pages), following:
$$\text{speedup} = \frac{k}{\rho}$$

{
We further compare the total GPU cost of \system, including training, with per-page LLM inference in Figure~\ref{fig:compute_comparison}. Let $g$ denote the number of groups processed. While per-page inference (dashed purple line) increases linearly with both the number of groups $g$ and the group size $k$, the \system-trained model yields substantial FLOP savings, with the magnitude of savings growing proportionally to group size. For instance, with $100$ pages per group, \system can already provide a computational saving of $1.12 \times 10^{21}$ FLOPS when processing $10^5$ groups. Additional details on the FLOP estimates are provided in Appendix~\ref{sec:appendix-compute-comparison}
}

Thus, compared to approaches that require per-page LLM inference ~\citep{bai2025autoschemakgautonomousknowledgegraph}, \system can significantly cut down the GPU resource usage for web-scale extraction.

\subsection{Ablations}
\label{sec:ablations}

\textbf{RQ3}: Does the \system reward design improve the model's capability in generating scripts that generalize to holdout elements?

To answer this question, we train a Q-14B model with the following reward for each training example $p$:
 \begin{equation}
 \label{eq:self_reward}
 r_{0}(p)= r_{\text{self}}(p)
 \end{equation}

Compared to Equation \ref{eq_scribes}, this reward encourages the model only to generate scripts suited to the current training example, without considering other in-group elements. We still use the same input prompt as in our \system-trained models (Prompt \ref{prompt:script-gen}), which instructs the model to produce scripts that generalize across similar webpages. The training setup remains unchanged.


As shown in Table \ref{tab:ablation_reward_design_q14b}, although this model outperforms Q-14B (\system) on the examples encountered during inference ($+1.2$\%), it generalizes much more poorly to similar webpages where the script is applied ($-7.2$\%), resulting in worse overall performance in the ``All'' column ($-4.2$\%). This shows that the \system reward design can more effectively instill in models the capability to produce generalizable scripts.

\begin{table}[t]
    \centering
    \footnotesize
    \resizebox{0.9\textwidth}{!}{
    \begin{tabular}{lccc|ccc|ccc}
    \toprule
    Model and Method & \multicolumn{3}{c}{All} & \multicolumn{3}{c}{Example}& \multicolumn{3}{c}{Holdout} \\
     \cmidrule(lr){2-4}\cmidrule(lr){5-7}\cmidrule(lr){8-10}
     & $R^{\mathrm{LM}}$ & $P^{\mathrm{LM}}$ & $F_1^{\mathrm{LM}}$ & $R^{\mathrm{LM}}$ & $P^{\mathrm{LM}}$ & $F_1^{\mathrm{LM}}$ & $R^{\mathrm{LM}}$ & $P^{\mathrm{LM}}$ & $F_1^{\mathrm{LM}}$ \\
    \midrule
    Q-14B (Reward w/ Eq. \ref{eq:self_reward})& 15.6 & 19.6 & 15.7 & 29.1 & \textbf{36.2} & \textbf{27.9} & 8.8 & 11.0 & 9.5 \\
    Q-14B (\system)& \textbf{23.0} & \textbf{24.3} & \textbf{19.9}  & \textbf{31.2} & 29.8 & 26.7 & \textbf{19.0} & \textbf{21.7}  & \textbf{16.7}\\
    \bottomrule
    \end{tabular}}
    \caption{Ablation study of reward design (Eq.~\ref{eq:self_reward}), showing that \system’s reward significantly enhances performance on holdout webpages.}
    \vspace{-5pt}
    \label{tab:ablation_reward_design_q14b}
    \end{table}

\begin{table}[t]
\centering
\footnotesize
\resizebox{0.9\textwidth}{!}{
\begin{tabular}{lccc|ccc|ccc}
\toprule
Method & \multicolumn{3}{c}{All}& \multicolumn{3}{c}{Example} & \multicolumn{3}{c}{Holdout} \\
 \cmidrule(lr){2-4}\cmidrule(lr){5-7}\cmidrule(lr){8-10}
 & $R^{\mathrm{LM}}$ & $P^{\mathrm{LM}}$ & $F_1^{\mathrm{LM}}$ & $R^{\mathrm{LM}}$ & $P^{\mathrm{LM}}$ & $F_1^{\mathrm{LM}}$ & $R^{\mathrm{LM}}$ & $P^{\mathrm{LM}}$ & $F_1^{\mathrm{LM}}$ \\
 \toprule
 Q-14B (Annotated only) & 23.0 & 24.3  & 19.9 & 31.2 & 29.8  & 26.7 & 19.0 & 21.7 & 16.7 \\
 Q-14B (+ All CC) & 22.0 & \textbf{30.2}  & \textbf{22.0} & 28.9 & 35.1  & 28.1 & 18.4 & \textbf{27.6} & \textbf{18.8} \\
 Q-14B (+ Failure-Case CC) & \textbf{25.2} & 23.0 & 21.8 & \textbf{34.9} & \textbf{31.0}  & \textbf{30.0} & \textbf{20.5} & 19.1 & 17.7 \\
\midrule

Q-32B (Annotated only)  & 29.9 & 31.5 & 28.1  & 32.0 & 33.9 & 30.3 & 28.8 & 30.3  & 26.8 \\
Q-32B (+ All CC)& 31.1 & 34.1  & 29.7 & 35.2 & \textbf{37.0} & 36.1 & 32.9 & 29.0 & 28.1 \\
Q-32B (+ Failure-Case CC) & \textbf{37.4} & \textbf{36.0} & \textbf{33.2} & \textbf{39.5} & 35.5 & \textbf{34.6}& \textbf{36.2} & \textbf{36.2}  & \textbf{32.4} \\

\bottomrule
\end{tabular}}
\caption{Ablation study on CC data subsets, showing that models trained with the failure-case subset generally perform better.}
\label{tab:cc_ablation_subsets}
\end{table}

\begin{table}[!t]
{
    \centering
    \footnotesize
    \resizebox{0.9\textwidth}{!}{
    \begin{tabular}{lccc|ccc|ccc}
    \toprule
    Model and Method & \multicolumn{3}{c}{All} & \multicolumn{3}{c}{Example}& \multicolumn{3}{c}{Holdout} \\
     \cmidrule(lr){2-4}\cmidrule(lr){5-7}\cmidrule(lr){8-10}
     & $R^{\mathrm{LM}}$ & $P^{\mathrm{LM}}$ & $F_1^{\mathrm{LM}}$ & $R^{\mathrm{LM}}$ & $P^{\mathrm{LM}}$ & $F_1^{\mathrm{LM}}$ & $R^{\mathrm{LM}}$ & $P^{\mathrm{LM}}$ & $F_1^{\mathrm{LM}}$ \\
    \midrule
    Q-14B agentic 3-iter 2-shot & 9.5 & 13.7 & 8.8 & 23.2 & 24.2 & 20.0 & 12.4 & 7.4 & 7.2 \\
    Q-14B (\system)& \textbf{20.7} & \textbf{22.2} & \textbf{19.4}  & \textbf{31.8} & \textbf{36.1} & \textbf{30.4} & \textbf{14.5} & \textbf{12.0}  & \textbf{12.2}\\
    \bottomrule
    \end{tabular}}
    \caption{{Ablation study on cross-domain transferability, showing that the \system-trained model demonstrate strong cross-domain transfer skills and outperform the baseline by more than 10\%.}}
    \vspace{-5pt}
    \label{tab:ablation_domain_seperation_q14b}}
    \end{table}

\textbf{RQ4}: Does using CommonCrawl data bring further improvements to our models?

We apply the technique described in Section~\ref{sec:reward_signal_unlabeled} to the final checkpoints of the \system-trained Q-14B and Q-32B models on the annotated dataset. As shown in Table~\ref{tab:llm_metrics_example_all}, additional training on synthetic data derived from CommonCrawl further improves performance, yielding gains of roughly 2\% for Q-14B and 5\% for Q-32B overall.


To better understand the impact of noisy rewards, we conducted the following ablation studies: (1) training directly on CC data, and (2) training on a mixture of CC and annotated data at a 1:1 ratio. Neither approach led to performance improvements, as shown in Table \ref{tab:cc_ablation_noisy} (Appendix \ref{sub:additional_ablation_noisy}). We therefore hypothesize that it is essential to first train the model with gold rewards to establish strong prior knowledge of this task. Subsequent training with noisy rewards can then expose the model to more diverse inputs, not only preserving but further improving performance, analogous to findings in \citet{shao2025spurious}.


\textbf{RQ5}: What's the effect of selecting the failure case subset to continue CommonCrawl trainings?

As discussed in Section \ref{sec:reward_signal_unlabeled}, we select the subset of CC data where our model produced scripts with no valid triples extracted. We examine whether restricting training to this subset is necessary by training both a 14B and a 32B model on the full CC dataset (``All CC'') and only the subset where no triples were extracted (``Failure-Case CC''). Results are reported in Table~\ref{tab:cc_ablation_subsets}. We highlight two findings: (1) Training on either All CC or Failure-Case CC improves performance compared to using annotated data alone, and (2) Failure-Case CC yields stronger gains for Q-32B compared to All CC (+3.5\%) , while performance for Q-14B remains comparable across the two settings.

{
\textbf{RQ6}: Do \system-trained models transfer across domains? For example, does a model trained on finance or legal tables generalize to product or encyclopedia pages?

To investigate this question, we conduct an ablation study using a train–test split in which the test set contains all product and encyclopedia pages, while the training set excludes webpages from these domains entirely. Details on this setup are provided in Appendix~\ref{sec:appendix-ablation-domain}. As shown in Table~\ref{tab:ablation_domain_seperation_q14b}, the \system-trained model still substantially outperforms the strongest agentic baseline of the same model by more than 10\%. To develop a model capable of web-scale extraction, we would still recommend training on a dataset that encompasses diverse domains and page layouts, as demonstrated by our CommonCrawl processing in Section~\ref{sec:reward_signal_unlabeled}.
}


\begin{figure}[t]
\centering

\begin{minipage}[t]{0.53\textwidth}
\vspace{0pt}
    \centering
    \rule{0pt}{0pt}
    \includegraphics[width=\linewidth]{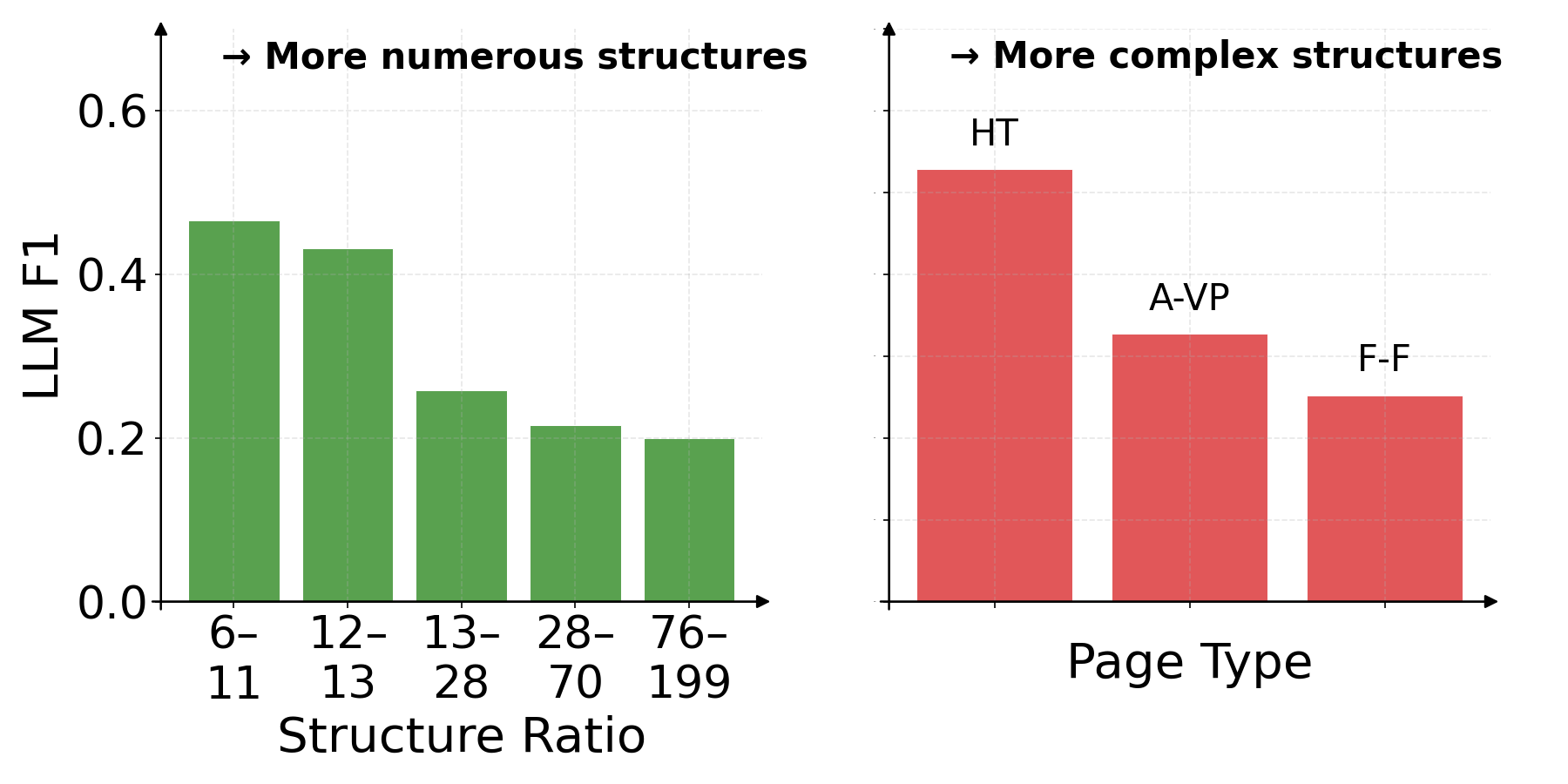}
    \subcaption{Performance of our best Q-32B model by amount of structure and page type.}
    \label{fig:error_analysis_a}
\end{minipage}
\hfill
\begin{minipage}[t]{0.43\textwidth}
\vspace{0pt}
    \centering
    \rule{0pt}{0pt}{
    \begin{tabular}{l c c c}
        \toprule
        Breakdown & $P^{LM}$ & $R^{LM}$ & $F_1^{LM}$ \\
        \midrule
        Nested list   & 23.6 & 18.7 & 19.7 \\
        Multi-column  & 23.3 & 33.9 & 18.6 \\
        All           & \textbf{39.5} & \textbf{35.5} & \textbf{34.6} \\
        \bottomrule
    \end{tabular}
    \subcaption{{Comparison of the best Q-32 model’s performance across nested-list and multi-column cases on our test set. \\
A nested list is defined as content containing a table or another list embedded within an outer list, while multi-column content refers to instances where headers span multiple columns.}}
    \label{fig:error_analysis_b}}
\end{minipage}

\caption{Error analysis: (a) model performance by structure complexity; {(b) comparison across nested lists and multi-column formats.}}
\label{fig:error_analysis_combined}

\end{figure}

\begin{table}[!t]
\small
\centering
\begin{tabular}{lcccccc}
\toprule
\textbf{Additional reference} & \textbf{Q-1.5B} & \textbf{Q-3B} & \textbf{Q-7B} & \textbf{Q-14B} & \textbf{Q-32B} & \textbf{GPT-4o} \\

\midrule


Flattened HTML & 50.2 & 53.8 & 62.9 & 74.2 & 70.8  & 82.5 \\

{ + Best Direct LLM Extraction triples} & {52.8} & {61.2} & {66.6} & {73.5} & {73.1} & {82.7}  \\
+ Best Q-32B triples & 52.9 & 54.3 & 64.1 & 77.3 & 73.2 & 86.6 \\
+ Ground truth triples & 60.5 & 64.9 & 70.5 & 78.2 & 74.8 & 87.4 \\

\bottomrule
\end{tabular}
\caption{QA accuracy (\%) with triple augmentations (evaluated by \texttt{Llama-3.3-instruct-70B}, Prompt~\ref{prompt:qa-eval}). \system’s predicted triples boost QA performance across many models.}
\label{tab:qa-performance}
\end{table}

\subsection{Error Analysis}

We perform an error analysis to understand the failures of the best-performing Q-32B model. We break down performance by the amount of structure in a webpage (approximated by the ratio of raw HTML length to flattened text length) and by webpage type. As shown on the left of Figure~\ref{fig:error_analysis_a} where webpages are grouped into five equal-sized bins (by number of webpages) and the respective medians are reported, performance declines as webpages contain more structure. On the right, the model performs best on webpages with Horizontal Tables (HT), followed by Attribute–Value Pairs (A-VP), and performs worst on Free-Form (F-F) pages. These results suggest that webpages with more numerous or complex structures are particularly challenging for our model. {

We also compare the performance of our model's outputs on contents involving multi-column and nested lists. As shown in Table \ref{fig:error_analysis_b}, we observe that such content is more challenging for our model. Further prediction examples are showcased in Appendix \ref{sec:appendix-example-predictions}.
}

\section{Downstream Applications}


\subsection{Question Answering over Semi-Structured Web Data}
\label{sec:qa-improvement}

We demonstrate that our script-extracted triples can enhance QA performance, even for the most capable LLMs. Although there exist many general-purpose QA datasets~\citep{yang2018hotpotqadatasetdiverseexplainable, rajpurkar2016squad100000questionsmachine} and datasets focused on semi-structured databases~\citep{chen-etal-2020-hybridqa, zhu-etal-2021-tat, chen2021openquestionansweringtables}, very few address the setting where the input consists of raw HTML. SemiBench~\citep{kai-2025-arxiv} fills this gap, containing QA pairs with aligned triple annotations.
This makes it a strong testbed for evaluating whether triple extraction improves QA over semi-structured web data. We select the subset of QA data (a total of 416 QA pairs) associated with our test set and evaluate a broad range of models as QA backbones, using the following reference conditions in Prompt~\ref{prompt:qa-reference}: (1) Flattened HTML only; {(2) Flattened HTML with best-performing direct LLM-extracted triples (GO-120B 2-shot flatten)}; (3) Flattened HTML with our model-extracted triples; and (4) Flattened HTML with gold triples. We report the result on the QA pairs associated with our validation examples in Table~\ref{tab:qa-performance}. Our \system-trained models yield consistent gains across diverse QA backbones, including an improvement of more than 4\% for GPT-4o.

{We further observe that although the \system-trained models slightly underperform the strongest per-page LLM-inference baseline in Table~\ref{tab:llm_metrics_example_all}, they nonetheless deliver comparable downstream QA gains. As shown in Table~\ref{tab:qa-performance}, using \system-generated triples improves QA performance for Q-14B and GPT-4o, yields roughly similar performance for Q-1.5B and Q-32B, and performs worse for Q-3B and Q-7B. These results indicate that higher IE accuracy does not necessarily translate into better downstream QA performance. Instead, using \system-produced triples can deliver much better efficiency and a similar level of downstream QA improvement.}


\subsection{Further Discussions}
The efficiency benefits of \system open up additional opportunities, and we highlight two directions for future explorations:  

\textbf{Multi-page, Complex QAs}: \system-extracted triples enable queries that require aggregation or ranking across multiple webpages. For example, a standard RAG solution would struggle with questions like ``What is the latest report filed?'' when answering against the website in Figure~\ref{fig2:similar_pages_demo}. In contrast, \system-generated triples can efficiently support such queries, eliminating the need for resource-intensive, page-by-page KG construction with LLMs.

\textbf{Pretraining}: Most open-source pretraining corpora systematically filter out semi-structured content. For instance, C4~\citep{raffel2023exploringlimitstransferlearning} applies a ``punctuation filter'' that removes sentences not ending with valid punctuation. Recent popular corpora such as Dolma~\citep{soldaini-etal-2024-dolma} and FineWeb~\citep{penedo2024finewebdatasetsdecantingweb} inherit this bias, resulting in a near-complete absence of semi-structured data. We believe \system can address this gap by enabling efficient and resource-effective extraction and incorporation of such content into pretraining datasets.

\section{Conclusion}

This work introduces a novel RL framework, \system, for training models to generate generalizable extraction scripts across structurally similar webpages for semi-structured content extraction. We also propose a new method for generating synthetic training data, which further improves model performance, by leveraging in-the-wild webpages from CommonCrawl. Experiments on our dataset demonstrate that \system-trained models yield substantial gains in question answering over semi-structured data. We hope that \system will facilitate further research on semi-structured content, such as complex QA and pretraining, and serve as a valuable tool for the community.


\bibliographystyle{iclr2026_conference}
\bibliography{iclr2026_conference}

@inproceedings{10.1145/2623330.2623623,
author = {Dong, Xin and Gabrilovich, Evgeniy and Heitz, Geremy and Horn, Wilko and Lao, Ni and Murphy, Kevin and Strohmann, Thomas and Sun, Shaohua and Zhang, Wei},
title = {Knowledge vault: a web-scale approach to probabilistic knowledge fusion},
year = {2014},
isbn = {9781450329569},
publisher = {Association for Computing Machinery},
address = {New York, NY, USA},
url = {https://doi.org/10.1145/2623330.2623623},
doi = {10.1145/2623330.2623623},
booktitle = {Proceedings of the 20th ACM SIGKDD International Conference on Knowledge Discovery and Data Mining},
pages = {601–610},
numpages = {10},
location = {New York, New York, USA},
series = {KDD '14}
}

@inproceedings{oguz-etal-2022-unik,
    title = "{U}ni{K}-{QA}: Unified Representations of Structured and Unstructured Knowledge for Open-Domain Question Answering",
    author = "Oguz, Barlas  and
      Chen, Xilun  and
      Karpukhin, Vladimir  and
      Peshterliev, Stan  and
      Okhonko, Dmytro  and
      Schlichtkrull, Michael  and
      Gupta, Sonal  and
      Mehdad, Yashar  and
      Yih, Scott",
    editor = "Carpuat, Marine  and
      de Marneffe, Marie-Catherine  and
      Meza Ruiz, Ivan Vladimir",
    booktitle = "Findings of the Association for Computational Linguistics: NAACL 2022",
    month = jul,
    year = "2022",
    address = "Seattle, United States",
    publisher = "Association for Computational Linguistics",
    url = "https://aclanthology.org/2022.findings-naacl.115/",
    doi = "10.18653/v1/2022.findings-naacl.115",
    pages = "1535--1546"
}

@inproceedings{ma-etal-2022-open,
    title = "Open Domain Question Answering with A Unified Knowledge Interface",
    author = "Ma, Kaixin  and
      Cheng, Hao  and
      Liu, Xiaodong  and
      Nyberg, Eric  and
      Gao, Jianfeng",
    editor = "Muresan, Smaranda  and
      Nakov, Preslav  and
      Villavicencio, Aline",
    booktitle = "Proceedings of the 60th Annual Meeting of the Association for Computational Linguistics (Volume 1: Long Papers)",
    month = may,
    year = "2022",
    address = "Dublin, Ireland",
    publisher = "Association for Computational Linguistics",
    url = "https://aclanthology.org/2022.acl-long.113/",
    doi = "10.18653/v1/2022.acl-long.113",
    pages = "1605--1620"
}

@inproceedings{gutiérrez2024hipporag,
      title={HippoRAG: Neurobiologically Inspired Long-Term Memory for Large Language Models}, 
      author={Bernal Jiménez Gutiérrez and Yiheng Shu and Yu Gu and Michihiro Yasunaga and Yu Su},
      booktitle={The Thirty-eighth Annual Conference on Neural Information Processing Systems},
      year={2024},
      url={https://openreview.net/forum?id=hkujvAPVsg}
}

@misc{penedo2024finewebdatasetsdecantingweb,
      title={The FineWeb Datasets: Decanting the Web for the Finest Text Data at Scale}, 
      author={Guilherme Penedo and Hynek Kydlíček and Loubna Ben allal and Anton Lozhkov and Margaret Mitchell and Colin Raffel and Leandro Von Werra and Thomas Wolf},
      year={2024},
      eprint={2406.17557},
      archivePrefix={arXiv},
      primaryClass={cs.CL},
      url={https://arxiv.org/abs/2406.17557}, 
}

@inproceedings{zhang-soh-2024-extract,
    title = "Extract, Define, Canonicalize: An {LLM}-based Framework for Knowledge Graph Construction",
    author = "Zhang, Bowen  and
      Soh, Harold",
    editor = "Al-Onaizan, Yaser  and
      Bansal, Mohit  and
      Chen, Yun-Nung",
    booktitle = "Proceedings of the 2024 Conference on Empirical Methods in Natural Language Processing",
    month = nov,
    year = "2024",
    address = "Miami, Florida, USA",
    publisher = "Association for Computational Linguistics",
    url = "https://aclanthology.org/2024.emnlp-main.548/",
    doi = "10.18653/v1/2024.emnlp-main.548",
    pages = "9820--9836"
}

@article{ning2023uukg,
  title={UUKG: Unified urban knowledge graph dataset for urban spatiotemporal prediction},
  author={Ning, Yansong and Liu, Hao and Wang, Hao and Zeng, Zhenyu and Xiong, Hui},
  journal={Advances in Neural Information Processing Systems},
  volume={36},
  pages={62442--62456},
  year={2023}
}

@INPROCEEDINGS{10386454,
  author={Chen, Bohan and Bertozzi, Andrea L.},
  booktitle={2023 IEEE International Conference on Big Data (BigData)}, 
  title={AutoKG: Efficient Automated Knowledge Graph Generation for Language Models}, 
  year={2023},
  volume={},
  number={},
  pages={3117-3126},
  doi={10.1109/BigData59044.2023.10386454}}

@inproceedings{zhang-etal-2023-aligning,
    title = "Aligning Instruction Tasks Unlocks Large Language Models as Zero-Shot Relation Extractors",
    author = "Zhang, Kai  and
      Jimenez Gutierrez, Bernal  and
      Su, Yu",
    editor = "Rogers, Anna  and
      Boyd-Graber, Jordan  and
      Okazaki, Naoaki",
    booktitle = "Findings of the Association for Computational Linguistics: ACL 2023",
    month = jul,
    year = "2023",
    address = "Toronto, Canada",
    publisher = "Association for Computational Linguistics",
    url = "https://aclanthology.org/2023.findings-acl.50/",
    doi = "10.18653/v1/2023.findings-acl.50",
    pages = "794--812"
}

@misc{wang2025readerlmv2smalllanguagemodel,
      title={ReaderLM-v2: Small Language Model for HTML to Markdown and JSON}, 
      author={Feng Wang and Zesheng Shi and Bo Wang and Nan Wang and Han Xiao},
      year={2025},
      eprint={2503.01151},
      archivePrefix={arXiv},
      primaryClass={cs.CL},
      url={https://arxiv.org/abs/2503.01151}, 
}

@misc{olmocr,
      title={{olmOCR: Unlocking Trillions of Tokens in PDFs with Vision Language Models}},
      author={Jake Poznanski and Jon Borchardt and Jason Dunkelberger and Regan Huff and Daniel Lin and Aman Rangapur and Christopher Wilhelm and Kyle Lo and Luca Soldaini},
      year={2025},
      eprint={2502.18443},
      archivePrefix={arXiv},
      primaryClass={cs.CL},
      url={https://arxiv.org/abs/2502.18443},
}

@inproceedings{barbaresi-2021-trafilatura,
  title = {{Trafilatura: A Web Scraping Library and Command-Line Tool for Text Discovery and Extraction}},
  author = "Barbaresi, Adrien",
  booktitle = "Proceedings of the Joint Conference of the 59th Annual Meeting of the Association for Computational Linguistics and the 11th International Joint Conference on Natural Language Processing: System Demonstrations",
  pages = "122--131",
  publisher = "Association for Computational Linguistics",
  url = "https://aclanthology.org/2021.acl-demo.15",
  year = 2021,
}

@inproceedings{zhang-etal-2024-spaghetti,
    title = "{SPAGHETTI}: Open-Domain Question Answering from Heterogeneous Data Sources with Retrieval and Semantic Parsing",
    author = "Zhang, Heidi  and
      Semnani, Sina  and
      Ghassemi, Farhad  and
      Xu, Jialiang  and
      Liu, Shicheng  and
      Lam, Monica",
    editor = "Ku, Lun-Wei  and
      Martins, Andre  and
      Srikumar, Vivek",
    booktitle = "Findings of the Association for Computational Linguistics: ACL 2024",
    month = aug,
    year = "2024",
    address = "Bangkok, Thailand",
    publisher = "Association for Computational Linguistics",
    url = "https://aclanthology.org/2024.findings-acl.96/",
    doi = "10.18653/v1/2024.findings-acl.96",
    pages = "1663--1678"
}

@inproceedings{10.1145/3477495.3531815,
author = {Christmann, Philipp and Saha Roy, Rishiraj and Weikum, Gerhard},
title = {Conversational Question Answering on Heterogeneous Sources},
year = {2022},
isbn = {9781450387323},
publisher = {Association for Computing Machinery},
address = {New York, NY, USA},
url = {https://doi.org/10.1145/3477495.3531815},
doi = {10.1145/3477495.3531815},
booktitle = {Proceedings of the 45th International ACM SIGIR Conference on Research and Development in Information Retrieval},
pages = {144–154},
numpages = {11},
location = {Madrid, Spain},
series = {SIGIR '22}
}

@misc{shao2024deepseekmathpushinglimitsmathematical,
      title={DeepSeekMath: Pushing the Limits of Mathematical Reasoning in Open Language Models}, 
      author={Zhihong Shao and Peiyi Wang and Qihao Zhu and Runxin Xu and Junxiao Song and Xiao Bi and Haowei Zhang and Mingchuan Zhang and Y. K. Li and Y. Wu and Daya Guo},
      year={2024},
      eprint={2402.03300},
      archivePrefix={arXiv},
      primaryClass={cs.CL},
      url={https://arxiv.org/abs/2402.03300}, 
}

@inproceedings{liu-etal-2024-spinach,
    title = "{SPINACH}: {SPARQL}-Based Information Navigation for Challenging Real-World Questions",
    author = "Liu, Shicheng  and
      Semnani, Sina  and
      Triedman, Harold  and
      Xu, Jialiang  and
      Zhao, Isaac Dan  and
      Lam, Monica",
    editor = "Al-Onaizan, Yaser  and
      Bansal, Mohit  and
      Chen, Yun-Nung",
    booktitle = "Findings of the Association for Computational Linguistics: EMNLP 2024",
    month = nov,
    year = "2024",
    address = "Miami, Florida, USA",
    publisher = "Association for Computational Linguistics",
    url = "https://aclanthology.org/2024.findings-emnlp.938/",
    doi = "10.18653/v1/2024.findings-emnlp.938",
    pages = "15977--16001"
}

@misc{kai-2025-arxiv,
  author       = {Kai Sun and Yin Huang and Srishti Mehra and Mohammad Kachuee and Xilun Chen and Renjie Tao and Zhaojiang Lin and Andrea Jessee and Nirav Shah and Alex Betty and Yue Liu and Anuj Kumar and Wen-tau Yih and Xin Luna Dong},
  title        = {Knowledge Extraction on Semi-Structured Content: Does It Remain Relevant for Question Answering in the Era of LLMs?},
  year         = {2025},
eprint={2509.25107},
      archivePrefix={arXiv},
      primaryClass={cs.CL},
      url={https://arxiv.org/abs/2509.25107},  
}

@misc{schulman2020klapprox,
  author       = {Schulman, Josh},
  title        = {Approximating KL Divergence},
  year         = {2020},
  howpublished = {Blog post},
}

@article{yao2022react,
  title={ReAct: Synergizing Reasoning and Acting in Language Models},
  author={Yao, Shunyu and Zhao, Jeffrey and Yu, Dian and Du, Nan and Shafran, Izhak and Narasimhan, Karthik and Cao, Yuan},
  journal={arXiv preprint arXiv:2210.03629},
  year={2022}
}

@inproceedings{soldaini-etal-2024-dolma,
    title = "Dolma: an Open Corpus of Three Trillion Tokens for Language Model Pretraining Research",
    author = "Soldaini, Luca  and
      Kinney, Rodney  and
      Bhagia, Akshita  and
      Schwenk, Dustin  and
      Atkinson, David  and
      Authur, Russell  and
      Bogin, Ben  and
      Chandu, Khyathi  and
      Dumas, Jennifer  and
      Elazar, Yanai  and
      Hofmann, Valentin  and
      Jha, Ananya  and
      Kumar, Sachin  and
      Lucy, Li  and
      Lyu, Xinxi  and
      Lambert, Nathan  and
      Magnusson, Ian  and
      Morrison, Jacob  and
      Muennighoff, Niklas  and
      Naik, Aakanksha  and
      Nam, Crystal  and
      Peters, Matthew  and
      Ravichander, Abhilasha  and
      Richardson, Kyle  and
      Shen, Zejiang  and
      Strubell, Emma  and
      Subramani, Nishant  and
      Tafjord, Oyvind  and
      Walsh, Evan  and
      Zettlemoyer, Luke  and
      Smith, Noah  and
      Hajishirzi, Hannaneh  and
      Beltagy, Iz  and
      Groeneveld, Dirk  and
      Dodge, Jesse  and
      Lo, Kyle",
    editor = "Ku, Lun-Wei  and
      Martins, Andre  and
      Srikumar, Vivek",
    booktitle = "Proceedings of the 62nd Annual Meeting of the Association for Computational Linguistics (Volume 1: Long Papers)",
    month = aug,
    year = "2024",
    address = "Bangkok, Thailand",
    publisher = "Association for Computational Linguistics",
    url = "https://aclanthology.org/2024.acl-long.840/",
    doi = "10.18653/v1/2024.acl-long.840",
    pages = "15725--15788"
}

@misc{raffel2023exploringlimitstransferlearning,
      title={Exploring the Limits of Transfer Learning with a Unified Text-to-Text Transformer}, 
      author={Colin Raffel and Noam Shazeer and Adam Roberts and Katherine Lee and Sharan Narang and Michael Matena and Yanqi Zhou and Wei Li and Peter J. Liu},
      year={2023},
      eprint={1910.10683},
      archivePrefix={arXiv},
      primaryClass={cs.LG},
      url={https://arxiv.org/abs/1910.10683}, 
}

@misc{firecrawl,
  author       = {Firecrawl},
  title        = {firecrawl: The Web Data API for AI – Turn entire websites into LLM-ready markdown or structured data},
  howpublished = {\url{https://github.com/firecrawl/firecrawl}},
  note         = {GitHub repository, licensed under AGPL-3.0, 54.3k stars, 4.6k forks (as of Sept 2 2025)},
  year         = {2025},
  month        = sep,
}

@misc{newspaper4k,
  author       = {Andrei Paraschiv},
  title        = {newspaper4k: Article Scraping \& Curation, a continuation of Newspaper3k},
  howpublished = {\url{https://github.com/AndyTheFactory/newspaper4k}},
  note         = {GitHub repository, a fork of Newspaper3k by codelucas; latest release v0.9.3 (March 18 2024), MIT license},
  year         = {2024},
  month        = mar,
}

@inproceedings{tan2024htmlrag,
author = {Tan, Jiejun and Dou, Zhicheng and Wang, Wen and Wang, Mang and Chen, Weipeng and Wen, Ji-Rong},
title = {HtmlRAG: HTML is Better Than Plain Text for Modeling Retrieved Knowledge in RAG Systems},
year = {2025},
isbn = {9798400712746},
publisher = {Association for Computing Machinery},
address = {New York, NY, USA},
url = {https://doi.org/10.1145/3696410.3714546},
doi = {10.1145/3696410.3714546},
booktitle = {Proceedings of the ACM on Web Conference 2025},
pages = {1733–1746},
numpages = {14},
location = {Sydney NSW, Australia},
series = {WWW '25}
}

@inproceedings{liu-etal-2024-suql,
    title = "{SUQL}: Conversational Search over Structured and Unstructured Data with Large Language Models",
    author = "Liu, Shicheng  and
      Xu, Jialiang  and
      Tjangnaka, Wesley  and
      Semnani, Sina  and
      Yu, Chen  and
      Lam, Monica",
    editor = "Duh, Kevin  and
      Gomez, Helena  and
      Bethard, Steven",
    booktitle = "Findings of the Association for Computational Linguistics: NAACL 2024",
    month = jun,
    year = "2024",
    address = "Mexico City, Mexico",
    publisher = "Association for Computational Linguistics",
    url = "https://aclanthology.org/2024.findings-naacl.283",
    pages = "4535--4555",
}

@inproceedings{10.5555/645927.672370,
author = {Crescenzi, Valter and Mecca, Giansalvatore and Merialdo, Paolo},
title = {RoadRunner: Towards Automatic Data Extraction from Large Web Sites},
year = {2001},
isbn = {1558608044},
publisher = {Morgan Kaufmann Publishers Inc.},
address = {San Francisco, CA, USA},
booktitle = {Proceedings of the 27th International Conference on Very Large Data Bases},
pages = {109–118},
numpages = {10},
series = {VLDB '01}
}

@inproceedings{10.1145/956750.956826,
author = {Liu, Bing and Grossman, Robert and Zhai, Yanhong},
title = {Mining data records in Web pages},
year = {2003},
isbn = {1581137370},
publisher = {Association for Computing Machinery},
address = {New York, NY, USA},
url = {https://doi.org/10.1145/956750.956826},
doi = {10.1145/956750.956826},
booktitle = {Proceedings of the Ninth ACM SIGKDD International Conference on Knowledge Discovery and Data Mining},
pages = {601–606},
numpages = {6},
location = {Washington, D.C.},
series = {KDD '03}
}

@inproceedings{10.1145/1060745.1060761,
author = {Zhai, Yanhong and Liu, Bing},
title = {Web data extraction based on partial tree alignment},
year = {2005},
isbn = {1595930469},
publisher = {Association for Computing Machinery},
address = {New York, NY, USA},
url = {https://doi.org/10.1145/1060745.1060761},
doi = {10.1145/1060745.1060761},
booktitle = {Proceedings of the 14th International Conference on World Wide Web},
pages = {76–85},
numpages = {10},
location = {Chiba, Japan},
series = {WWW '05}
}

@misc{dalvi2011automatic,
    title={Automatic Wrappers for Large Scale Web Extraction},
    author={Nilesh Dalvi and Ravi Kumar and Mohamed Soliman},
    year={2011},
    eprint={1103.2406},
    archivePrefix={arXiv},
    primaryClass={cs.DB}
}

@misc{commoncrawl,
  author       = {{Common Crawl}},
  title        = {Common Crawl},
  howpublished = {\url{https://commoncrawl.org/}},
  note         = {Accessed: 2025-08},
  year         = {2025}
}

@misc{peng2023yarn,
    title={YaRN: Efficient Context Window Extension of Large Language Models},
    author={Bowen Peng and Jeffrey Quesnelle and Honglu Fan and Enrico Shippole},
    year={2023},
    eprint={2309.00071},
    archivePrefix={arXiv},
    primaryClass={cs.CL}
}

@misc{zhao2023pytorchfsdpexperiencesscaling,
      title={PyTorch FSDP: Experiences on Scaling Fully Sharded Data Parallel}, 
      author={Yanli Zhao and Andrew Gu and Rohan Varma and Liang Luo and Chien-Chin Huang and Min Xu and Less Wright and Hamid Shojanazeri and Myle Ott and Sam Shleifer and Alban Desmaison and Can Balioglu and Pritam Damania and Bernard Nguyen and Geeta Chauhan and Yuchen Hao and Ajit Mathews and Shen Li},
      year={2023},
      eprint={2304.11277},
      archivePrefix={arXiv},
      primaryClass={cs.DC},
      url={https://arxiv.org/abs/2304.11277}, 
}

@inproceedings{kushmerick1997wrapper,
  title={Wrapper induction for information extraction},
  author={Kushmerick, Nicholas and Weld, Daniel S and Doorenbos, Robert B},
  booktitle={Proceedings of the Fifteenth International Joint Conference on Artificial Intelligence (IJCAI)},
  pages={729--737},
  year={1997}
}

@inproceedings{freitag2000boosted,
  title={Boosted wrapper induction},
  author={Freitag, Dayne and Kushmerick, Nicholas},
  booktitle={Proceedings of the AAAI Conference on Artificial Intelligence},
  pages={577--583},
  year={2000}
}

@misc{bai2025autoschemakgautonomousknowledgegraph,
      title={AutoSchemaKG: Autonomous Knowledge Graph Construction through Dynamic Schema Induction from Web-Scale Corpora}, 
      author={Jiaxin Bai and Wei Fan and Qi Hu and Qing Zong and Chunyang Li and Hong Ting Tsang and Hongyu Luo and Yauwai Yim and Haoyu Huang and Xiao Zhou and Feng Qin and Tianshi Zheng and Xi Peng and Xin Yao and Huiwen Yang and Leijie Wu and Yi Ji and Gong Zhang and Renhai Chen and Yangqiu Song},
      year={2025},
      eprint={2505.23628},
      archivePrefix={arXiv},
      primaryClass={cs.CL},
      url={https://arxiv.org/abs/2505.23628}, 
}

@misc{yang2018hotpotqadatasetdiverseexplainable,
      title={HotpotQA: A Dataset for Diverse, Explainable Multi-hop Question Answering}, 
      author={Zhilin Yang and Peng Qi and Saizheng Zhang and Yoshua Bengio and William W. Cohen and Ruslan Salakhutdinov and Christopher D. Manning},
      year={2018},
      eprint={1809.09600},
      archivePrefix={arXiv},
      primaryClass={cs.CL},
      url={https://arxiv.org/abs/1809.09600}, 
}

@misc{rajpurkar2016squad100000questionsmachine,
      title={SQuAD: 100,000+ Questions for Machine Comprehension of Text}, 
      author={Pranav Rajpurkar and Jian Zhang and Konstantin Lopyrev and Percy Liang},
      year={2016},
      eprint={1606.05250},
      archivePrefix={arXiv},
      primaryClass={cs.CL},
      url={https://arxiv.org/abs/1606.05250}, 
}

@inproceedings{chen-etal-2020-hybridqa,
    title = "{H}ybrid{QA}: A Dataset of Multi-Hop Question Answering over Tabular and Textual Data",
    author = "Chen, Wenhu  and
      Zha, Hanwen  and
      Chen, Zhiyu  and
      Xiong, Wenhan  and
      Wang, Hong  and
      Wang, William Yang",
    editor = "Cohn, Trevor  and
      He, Yulan  and
      Liu, Yang",
    booktitle = "Findings of the Association for Computational Linguistics: EMNLP 2020",
    month = nov,
    year = "2020",
    address = "Online",
    publisher = "Association for Computational Linguistics",
    url = "https://aclanthology.org/2020.findings-emnlp.91/",
    doi = "10.18653/v1/2020.findings-emnlp.91",
    pages = "1026--1036"
}

@inproceedings{zhu-etal-2021-tat,
    title = "{TAT}-{QA}: A Question Answering Benchmark on a Hybrid of Tabular and Textual Content in Finance",
    author = "Zhu, Fengbin  and
      Lei, Wenqiang  and
      Huang, Youcheng  and
      Wang, Chao  and
      Zhang, Shuo  and
      Lv, Jiancheng  and
      Feng, Fuli  and
      Chua, Tat-Seng",
    editor = "Zong, Chengqing  and
      Xia, Fei  and
      Li, Wenjie  and
      Navigli, Roberto",
    booktitle = "Proceedings of the 59th Annual Meeting of the Association for Computational Linguistics and the 11th International Joint Conference on Natural Language Processing (Volume 1: Long Papers)",
    month = aug,
    year = "2021",
    address = "Online",
    publisher = "Association for Computational Linguistics",
    url = "https://aclanthology.org/2021.acl-long.254/",
    doi = "10.18653/v1/2021.acl-long.254",
    pages = "3277--3287"
}

@misc{chen2021openquestionansweringtables,
      title={Open Question Answering over Tables and Text}, 
      author={Wenhu Chen and Ming-Wei Chang and Eva Schlinger and William Wang and William W. Cohen},
      year={2021},
      eprint={2010.10439},
      archivePrefix={arXiv},
      primaryClass={cs.CL},
      url={https://arxiv.org/abs/2010.10439}, 
}

@article{10.14778/3231751.3231758,
author = {Lockard, Colin and Dong, Xin Luna and Einolghozati, Arash and Shiralkar, Prashant},
title = {CERES: distantly supervised relation extraction from the semi-structured web},
year = {2018},
issue_date = {June 2018},
publisher = {VLDB Endowment},
volume = {11},
number = {10},
issn = {2150-8097},
url = {https://doi.org/10.14778/3231751.3231758},
doi = {10.14778/3231751.3231758},
journal = {Proc. VLDB Endow.},
month = jun,
pages = {1084–1096},
numpages = {13}
}

@inproceedings{lockard-etal-2020-zeroshotceres,
    title = "{Z}ero{S}hot{C}eres: Zero-Shot Relation Extraction from Semi-Structured Webpages",
    author = "Lockard, Colin  and
      Shiralkar, Prashant  and
      Dong, Xin Luna  and
      Hajishirzi, Hannaneh",
    editor = "Jurafsky, Dan  and
      Chai, Joyce  and
      Schluter, Natalie  and
      Tetreault, Joel",
    booktitle = "Proceedings of the 58th Annual Meeting of the Association for Computational Linguistics",
    month = jul,
    year = "2020",
    address = "Online",
    publisher = "Association for Computational Linguistics",
    url = "https://aclanthology.org/2020.acl-main.721/",
    doi = "10.18653/v1/2020.acl-main.721",
    pages = "8105--8117"
}

@article{zuo2025ttrl,
  title={Ttrl: Test-time reinforcement learning},
  author={Zuo, Yuxin and Zhang, Kaiyan and Sheng, Li and Qu, Shang and Cui, Ganqu and Zhu, Xuekai and Li, Haozhan and Zhang, Yuchen and Long, Xinwei and Hua, Ermo and others},
  journal={arXiv preprint arXiv:2504.16084},
  year={2025}
}

@article{zhao2025learning,
  title={Learning to reason without external rewards},
  author={Zhao, Xuandong and Kang, Zhewei and Feng, Aosong and Levine, Sergey and Song, Dawn},
  journal={arXiv preprint arXiv:2505.19590},
  year={2025}
}

@article{shao2025spurious,
  title={Spurious rewards: Rethinking training signals in rlvr},
  author={Shao, Rulin and Li, Shuyue Stella and Xin, Rui and Geng, Scott and Wang, Yiping and Oh, Sewoong and Du, Simon Shaolei and Lambert, Nathan and Min, Sewon and Krishna, Ranjay and others},
  journal={arXiv preprint arXiv:2506.10947},
  year={2025}
}

@article{prabhudesai2025maximizing,
  title={Maximizing Confidence Alone Improves Reasoning},
  author={Prabhudesai, Mihir and Chen, Lili and Ippoliti, Alex and Fragkiadaki, Katerina and Liu, Hao and Pathak, Deepak},
  journal={arXiv preprint arXiv:2505.22660},
  year={2025}
}

@inproceedings{10.5555/645856.758279,
author = {Wilks, Yorick},
title = {Information Extraction as a Core Language Technology},
year = {1997},
isbn = {354063438X},
publisher = {Springer-Verlag},
address = {Berlin, Heidelberg},
booktitle = {International Summer School on Information Extraction: A Multidisciplinary Approach to an Emerging Information Technology},
pages = {1–9},
numpages = {9},
series = {SCIE '97}
}

\appendix

\newpage
\appendix
\section*{Appendix}
\startcontents[sections]
\printcontents[sections]{l}{1}{\setcounter{tocdepth}{2}}
\newpage

\section{Use of LLMs in this Research}

We utilize LLMs in two main ways in this research:

\begin{enumerate}
    \item \textbf{Assistance with Code Writing:} During the implementation of RL training and evaluation scripts, LLMs were occasionally used as assistants. All code was subsequently double-checked and verified by the authors.
    \item \textbf{Paper Language and Related Works:} During the writing process, we occasionally utilized LLMs to improve the clarity and fluency of the English. We also occasionally use LLM-assisted search systems to find additional related works. All final text was reviewed by the authors.
\end{enumerate}

\section{Websites with Semi-Structured Content}
\label{sec:semi-structured_content_appendix}


We can broadly classify webpages with semi-structured content into three categories:
\begin{enumerate}
\item \textbf{Horizontal Tables}: These webpages primarily present information in a tabular format.
\item \textbf{Attribute-Value Pairs}: Information is organized as attribute-value pairs, typically displayed across multiple rows in an “infobox”-like format.
\item \textbf{Free Form}: Semi-structured content is distributed throughout the page, often combining both horizontal tables and attribute-value pairs.
\end{enumerate}

For additional information and more details on these breakdowns, refer to \citet{kai-2025-arxiv}.


\section{HTML Dedup Algorithm Details}
\label{sec:appendix-dedup-algo}

\begin{figure}[ht]
    \centering
    \includegraphics[width=\textwidth]{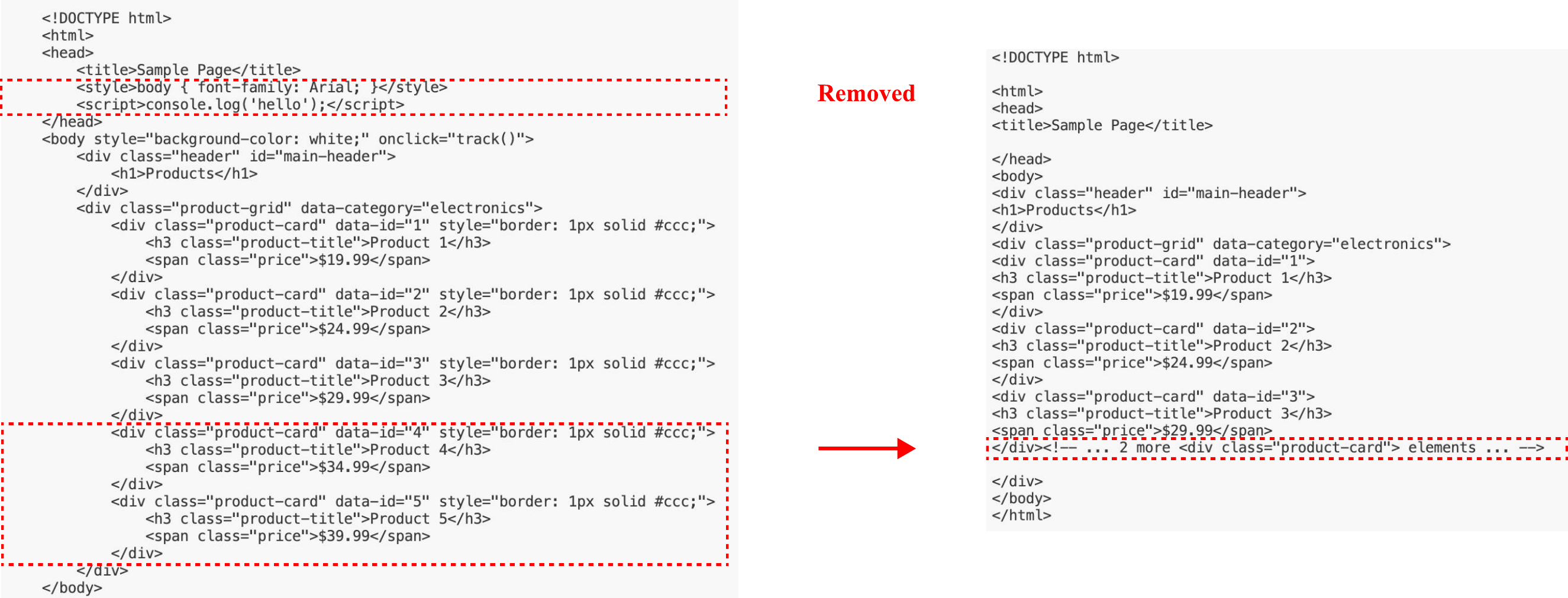}
    \caption{An example illustrating Algorithm~\ref{alg:dedup} is shown here. The original HTML appears on the left, while the compressed HTML is shown on the right. The dashed-highlighted section near the top, containing script and style elements, has been removed. The repeated HTML content near the bottom has been deduplicated, retaining up to $z=3$ elements.}
    \label{fig:dedup_example}
\end{figure}

Raw HTMLs are often long and repetitive. We propose a simple and effective dedup algorithm to significantly cut down the token length of HTML pages while still maintaining its structure. Algorithm \ref{alg:dedup} shows the implementation of this algorithm. We set $z=3$ in our experiments.

Table \ref{tab:token_reduction} shows the token saving effect of our dedup algorithm. Removing whitespaces in a HTML only brings minimal token savings ($<$ 2\%), while our dedup algorithm brings significant token savings, cutting down token usage from $>$114k to $<$17k. We also profiled performance gains of baselines models using dedup. As shown in Table \ref{tab:dedup_ablation}, employing deduplicated HTML yields clear improvements compared to using raw HTML. Most notably, deduplication significantly increases the Non-Empty Rate of baseline performance by enabling more data points to fit within the model's context window.

\begin{algorithm}[ht]
\caption{Structure-Preserving HTML Deduplication (keep-$z$)}
\begin{algorithmic}[1]
\Require Raw HTML string $H$, integer $z \ge 1$ (default $z{=}3$)
\Ensure Compressed, structure-preserving HTML
\State Parse $H$ into DOM $R$ (fallback parser if needed; return $H$ on failure)
\State $\mathrm{RemoveTags} \gets \{\texttt{script}, \texttt{style}, \texttt{noscript},$
\Statex \hspace{\algorithmicindent} $\texttt{iframe}, \texttt{embed}, \texttt{object}, \texttt{applet},$
\Statex \hspace{\algorithmicindent} $\texttt{meta}, \texttt{link}, \texttt{base}\}$
\State $\mathrm{KeepAttrs} \gets \{\texttt{id}, \texttt{class}, \texttt{role}, \texttt{name},$
\Statex \hspace{\algorithmicindent} $\texttt{type}, \texttt{href}, \texttt{src}, \texttt{alt}, \texttt{title},$
\Statex \hspace{\algorithmicindent} $\texttt{rel}, \texttt{target}, \texttt{for}, \texttt{action}, \texttt{method},$
\Statex \hspace{\algorithmicindent} $\texttt{value}, \texttt{placeholder}, \texttt{required}, \texttt{data-*}, \texttt{aria-*}\}$
\State Remove all nodes with tag in $\mathrm{RemoveTags}$
\State Remove all HTML comments except those starting with ``\texttt{...}''
\ForAll{element nodes $e$ in $R$}
  \ForAll{attributes $a$ of $e$}
    \If{$a \notin \mathrm{KeepAttrs}$ and $a$ not prefixed by \texttt{data-} or \texttt{aria-}}
      \State delete attribute $a$ from $e$
    \EndIf
  \EndFor
\EndFor
\ForAll{nodes $n$ in traversal of $R$}
  \If{$n.\mathrm{tag} \in \{\texttt{ul}, \texttt{ol}, \texttt{div}, \texttt{section}, \texttt{tbody}, \texttt{thead}, \texttt{select}\}$}
    \State $\mathrm{children} \gets [\, c \in n.\mathrm{children} : c \text{ is an element}\,]$
    \State Group $\mathrm{children}$ by $\mathrm{sig}(c) \gets (c.\mathrm{tag}, \mathrm{sort}(c.\mathrm{class}\ \text{or}\ [\,]))$
    \ForAll{group $G$}
      \If{$|G| > z$}
        \State Keep the first $z$ in $G$ (order preserved); remove the rest
        \State After the $z$-th kept node, insert comment:
        \State \quad ``\texttt{ ... $|G|-z$ more <$\mathrm{tag}$ class='...'> elements ... }''
      \EndIf
    \EndFor
  \EndIf
\EndFor
\State Optionally normalize whitespace and excessive blank lines
\State \Return serialized DOM
\end{algorithmic}
\label{alg:dedup}
\end{algorithm}

\begin{table}[ht]
\centering
\begin{tabular}{lrr}
\hline
\textbf{Processing Stage} & \textbf{Avg Tokens} & \textbf{Percentage} \\
\hline
Original tokens & 114,318.6 & 100.0\% \\
After whitespace removal & 112,279.0 & 98.2\% \\
After dedup & 16,985.1 & 14.9\% \\
\hline
\multicolumn{3}{l}{\textbf{Reductions}} \\
Whitespace token savings & 2,039.6 & 1.8\% \\
Total dedup token savings & 97,333.5 & 85.1\% \\
\hline
\end{tabular}
\caption{Token reduction analysis across the webpages collected by \citet{kai-2025-arxiv}. Tokens were profiled with GPT-4o tokenizer, accessed via \url{https://github.com/openai/tiktoken}.}
\label{tab:token_reduction}
\end{table}

\begin{table}[ht]
    \centering
    \small
    \renewcommand{\arraystretch}{1.2}
    \begin{tabular}{@{}lcccc@{}}
        \toprule
        \textbf{Model \& Format} & $P^{\text{LM}}$ & $R^{\text{LM}}$ & $F_1^{\text{H, LM}}$ & Non-Empty Rate  \\
        \midrule
        L-70B w/ Raw HTML & 3.4 & 3.7 & 3.5 & 37.9 \\
        L-70B w/ Dedup HTML & 14.2 & 9.5 & 11.3 & 46.4 \\
        \midrule
        GPT-4o w/ Raw HTML & 13.7 & 15.4 & 14.5 & 63.8 \\
        GPT-4o w/ Dedup HTML & 19.1 & 23.0 & 20.9 & 94.9 \\
        \bottomrule
    \end{tabular}
    \caption{Performance comparison of baseline models using raw or dedup-ed HTML. Here, we feed each page in one-by-one in this dataset and only evaluate the model's performance on one given page. Non-Empty Rate is set to $1$ if the model's generated code produced at least 1 triple on this page, and $0$ if otherwise.}
    \label{tab:dedup_ablation}
\end{table}

\section{Training Hyperparameters and Other Details}
\label{sec:appendix-training-setup}

\subsection{Data Pre-processing}
\label{sec:appendix-data-proprocessing}
During training, we set the maximum prompt length to $28672$ tokens and the maximum response length to $4096$ tokens. This results in a total model context window of $32768$ tokens, which is the maximum length before needing to apply YaRN~\citep{peng2023yarn} for the Qwen-2.5 series models\footnote{We observed empirically that model training with YaRN becomes much more unstable and difficult to converge.}.

SemiBench~\citep{kai-2025-arxiv} includes a subset of 268 webpages drawn from 56 groups, each containing more than one webpage. We partition the groups into training and test sets at an approximately 6:4 ratio, resulting in 34 groups (192 webpages) for training and 22 groups (76 webpages) for testing. After applying the maximum-context constraint described above, 141 training webpages and 65 test webpages remain.


\subsection{Training Details}
\label{sec:appendix-training-details}

During GRPO training, we do not apply entropy loss. We set the KL loss coefficient to $0.001$ and the KL loss to be the $k_3$ loss using the approximation described in \citet{schulman2020klapprox}, i.e., 
\[
k_{3}(a) = \frac{\pi_{\text{new}}(a)}{\pi_{\text{old}}(a)} \;-\; 
\log \frac{\pi_{\text{new}}(a)}{\pi_{\text{old}}(a)} \;-\; 1
\]

We use the default model rollout parameters (for Qwen-2.5-instruct, these are top\_k$ =-1$, top\_p$ =1$, and temperature $=1$) and validation/inference parameters (for Qwen-2.5-instruct, these are top\_k$ =-1$, top\_p$ =1$, and temperature $=0$). We do not use LoRA and instead perform full-parameter finetuning with FSDP~\citep{zhao2023pytorchfsdpexperiencesscaling}. We trained the models on the annotated set for a total of 50 epochs, and on CommonCrawl data for 1 epoch. For each update, we collect 8 rollouts to perform GRPO update. For the 32B model, we apply a $0.5$ gradient clipping, which we found to lead to more stable trainings. We set the learning rate to be a constant $1e-6$.

{
\subsection{Detailed on Compute Comparison}
\label{sec:appendix-compute-comparison}

To train the Q-14B \system model, we ran approximately 12 hours of training at an average throughput of 3,958 TFLOPs across all GPUs, yielding an estimated total training compute of $1.71 \times 10^{20}$ FLOPs. For the Q-32B \system model, we similarly trained for about 12 hours at an average of 8,232 TFLOPs, resulting in an estimated total compute of $3.56 \times 10^{20}$ FLOPs.

To estimate the per-page inference cost reported in Figure~\ref{fig:compute_comparison}, we assume that each forward pass requires roughly 2 times the parameter count per token.
}

{
\section{Dataset Comparison}

We compare several statistics of HTML webpages in Table~\ref{tab:html_feature_stats_all}. Below, we define each statistic:

\textbf{DOM Max Depth}: The maximum depth of the Document Object Model (DOM) tree in an HTML document. This measures how deeply the elements are nested; a higher DOM Max Depth indicates more extensive nesting.

\textbf{Deduplication Ratio}: The lengths of the HTML content before and after applying the deduplication algorithm described in Appendix~\ref{sec:appendix-dedup-algo} (in characters). This quantifies redundancy in the HTML structure; a lower Deduplication Ratio indicates greater redundancy.

\textbf{Structure Ratio}: The ratio of the HTML length to the flattened text length (in characters). This approximates how much structural markup the HTML contains relative to its textual content; a higher Structure Ratio reflects more structural complexity.

\textbf{Tag Count}: The number of all tags in an HTML document\footnote{\texttt{len(soup.find\_all(True))}}. This measures the structural complexity of the HTML; a higher Tag Count indicates a more complex document.
}


\begin{table}[ht]
  \centering
  {
  \begin{tabular}{l l r r r}
    \toprule
    Feature & Metric & Train & Test & CC (After Step 6 in Fig.~\ref{fig:commoncrawl}) \\
    \midrule
    \multirow{5}{*}{DOM Max Depth} & Mean & 20.2 & 18.4 & 20.2 \\
     & Median & 19.0 & 17.0 & 17.0 \\
     & Std & 4.94 & 6.98 & 21.7 \\
     & Min & 10.0 & 10.0 & 5.00 \\
     & Max & 37.0 & 37.0 & 455 \\
    \midrule
    \multirow{5}{*}{Deduplication Ratio} & Mean & 0.215 & 0.174 & 0.353 \\
     & Median & 0.213 & 0.166 & 0.344 \\
     & Std & 0.111 & 0.0986 & 0.178 \\
     & Min & 0.0302 & 0.0324 & 0.000480 \\
     & Max & 0.484 & 0.422 & 1.02 \\
    \midrule
    \multirow{5}{*}{Structure Ratio} & Mean & 46.1 & 43.5 & 20.8 \\
     & Median & 28.7 & 27.4 & 13.8 \\
     & Std & 40.0 & 45.5 & 48.7 \\
     & Min & 2.24 & 6.00 & 1.19 \\
     & Max & 174 & 199 & 1960 \\
    \midrule
    \multirow{5}{*}{Tag Count} & Mean & 1650 & 1820 & 655 \\
     & Median & 1260 & 1080 & 496 \\
     & Std & 2140 & 2550 & 559 \\
     & Min & 224 & 154 & 18.0 \\
     & Max & 27800 & 12300 & 5070 \\
    \bottomrule
  \end{tabular}
  \caption{{Summary statistics (Mean, Median, Std, Min, Max) for HTML-derived features across datasets.}}
  \label{tab:html_feature_stats_all}}
\end{table}

{

This comparison shows that the labeled training and test sets share similar summary statistics, whereas the CommonCrawl portion differs noticeably. In particular, the CommonCrawl data is less redundant (lower Deduplication Ratio), contains less structural markup (lower Structure Ratio), and is structurally simpler (lower Tag Count). Across all metrics, it also exhibits greater variability, as indicated by the higher standard deviations. These observations suggest that incorporating this portion of the CommonCrawl data into training can meaningfully broaden the distribution of inputs, exposing the models to examples that differ substantially from those in the labeled dataset.
}

\section{Metrics and their implementation}
\label{sub:appendix_metrics}



\subsection{Details on the Fuzzy Match Algorithm}
\label{sub:appendix_metrics_details}

Formally, let $G = \{g_1, g_2, \dots, g_m\}$ denote the set of gold triples and $P = \{p_1, p_2, \dots, p_n\}$ the predicted triples. Instead of requiring exact equality, we define a similarity function $f^{\text{fuzzy}}(g_i, p_j) \in [0,1]$ that quantifies the degree of match between a gold triple $g_i$ and a predicted triple $p_j$ as the ratio of character-level matching\footnote{Implemented via \url{https://github.com/seatgeek/fuzzywuzzy}'s ratio function, which calculate a ratio of character-level matching using \href{https://en.wikipedia.org/wiki/Levenshtein_distance}{Levenshtein distance} .}. To ensure one-to-one alignment, we compute a maximum-weight bipartite matching between $G$ and $P$, where the weight of each edge is $f^{\text{fuzzy}}(g_i, p_j)$. This assignment is efficiently solved using the Jonker--Volgenant algorithm\footnote{Implemented via \url{https://docs.scipy.org/doc/scipy/reference/generated/scipy.optimize.linear_sum_assignment.html}.}. Precision, recall, and $F_1$ are then generalized as:
\[
P^{\text{fuzzy}} = \frac{\sum_{(g,p) \in M} f^{\text{fuzzy}}(g,p)}{|P|}, \quad 
R^{\text{fuzzy}} = \frac{\sum_{(g,p) \in M} f^{\text{fuzzy}}(g,p)}{|G|}, \quad 
F_1^{\text{fuzzy}} = \frac{2 \cdot P^{\text{fuzzy}} \cdot R^{\text{fuzzy}}}{P^{\text{fuzzy}} + R^{\text{fuzzy}}}.
\]

where $M \subseteq G \times P$ denotes the optimal matching. Given $M$, the LLM-based metric evaluates correctness by invoking a LLM on the final matched pairs of gold and predicted triples. For each pair $(g,p) \in M$, the model outputs a binary judgment $f^{\text{LM}}(g,p) \in \{0,1\}$, where $1$ denotes a true match and $0$ denotes a failed match according to Prompt \ref{prompt:llm-as-a-judge}. We then define LLM-based precision, recall, and $F_1$ as:
\[
P^{\text{LM}} = \frac{\sum_{(g,p) \in M} f^{\text{LM}}(g,p)}{|P|}, \quad
R^{\text{LM}} = \frac{\sum_{(g,p) \in M} f^{\text{LM}}(g,p)}{|G|}, \quad
F_1^{\text{LM}} = \frac{2 \cdot P^{\text{LM}} \cdot R^{\text{LM}}}{P^{\text{LM}} + R^{\text{LM}}}.
\]

{
Empirically, we observe a correlation between $F_1^{\text{fuzzy}}$ and $F_1^{\text{LM}}$. The latter tends to yield slightly lower absolute scores but exhibits the same performance trend across models and configurations. A comparison showing the two metrics and the associated precision and recall metrics for the baselines are shown in Table \ref{tab:llm_vs_fuzzy_all}. We calculated the correlation coefficient between $F_1^{\text{fuzzy}}$ and $F_1^{\text{LM}}$ to be $0.957$ with a p-value of $1.4 \times 10^{-5}$, showing a strong positive correlation.

\begin{table}[t]
\centering
\small
{
\begin{tabular}{lcccccc}
\toprule
Method & $R^{\mathrm{LM}}$ & $P^{\mathrm{LM}}$ & \cellcolor{gray!10}$F_1^{\mathrm{LM}}$ & $R^{\mathrm{fuzzy}}$ & $P^{\mathrm{fuzzy}}$ & \cellcolor{gray!10}$F_1^{\mathrm{fuzzy}}$ \\
\midrule
Q-14B flatten & 30.46 & 36.46 & \cellcolor{gray!10} 29.87 & 45.96 & 52.37 & \cellcolor{gray!10} 43.50 \\
Q-32B flatten & 28.73 & 37.44 & \cellcolor{gray!10} 29.93 & 41.62 & 54.25 & \cellcolor{gray!10} 42.26 \\
GO-20B 2-shot flatten & 33.18 & 47.10 & \cellcolor{gray!10} 34.93 & 46.53 & 65.21 & \cellcolor{gray!10} 49.77 \\
GO-120B 2-shot flatten & 42.27 & 46.26 & \cellcolor{gray!10} 40.40 & 56.01 & 61.42 & \cellcolor{gray!10} 53.37 \\

Q-14B  3-iter 2-shot & 8.59 & 11.13 & \cellcolor{gray!10} 8.01 & 17.17 & 25.57 & \cellcolor{gray!10} 16.53 \\
Q-72B  3-iter 2-shot & 16.40 & 19.41 & \cellcolor{gray!10} 14.97 & 28.73 & 37.96 & \cellcolor{gray!10} 28.60 \\
Q-32B  3-iter 2-shot & 18.56 & 27.20 & \cellcolor{gray!10} 19.41 & 27.49 & 44.67 & \cellcolor{gray!10} 30.39 \\
GO-20B  3-iter & 24.70 & 23.22 & \cellcolor{gray!10} 20.87 & 52.30 & 41.83 & \cellcolor{gray!10} 39.58 \\
GPT-4o  3-iter 2-shot & 25.95 & 33.04 & \cellcolor{gray!10} 24.42 & 45.58 & 60.57 & \cellcolor{gray!10} 44.46 \\
GO-120B  3-iter 2-shot & 33.86 & 40.96 & \cellcolor{gray!10} 34.30 & 49.79 & 65.72 & \cellcolor{gray!10} 52.02 \\
\bottomrule
\end{tabular}
\caption{{Comparison of LLM-judged metrics and fuzzy-matching metrics for baselines reported in Table \ref{tab:llm_metrics_example_all} for the ``All'' column. Gray-highlighted columns denote $F_1^{\mathrm{LM}}$ and $F_1^{\mathrm{fuzzy}}$. This comparison shows that the two metrics show similar performance trend across models and configurations.}}
\label{tab:llm_vs_fuzzy_all}
}
\end{table}

}

\subsection{Reward during RL implementation}
\label{sub:appendix_reward_during_RL}

We use $F_1^{\text{fuzzy}}$ during training as a proxy for $F_1^{\text{LM}}$, thereby avoiding LLM calls. Because computing fuzzy $F_1$ exactly requires solving a maximum-weight bipartite matching, runtime can become too long for large sets of triples. We thus approximate the matching with a greedy heuristic. Specifically, all candidate pairs of gold and predicted triples are scored by $f^{\text{fuzzy}}$, sorted in descending order, and added sequentially to the matching as long as they do not conflict with previously chosen pairs. This yields a fast, albeit sub-optimal, alignment. To ensure scalability, we impose a 60-seconds cutoff for evaluation. If timeout occurs, we further project the total similarity score by extrapolating from the average score of observed matches to the remaining unmatched capacity.

{
\subsection{Human verification of RL reward}

We followed the same evaluation metrics as defined in \citet{kai-2025-arxiv}, which reported a 95\% agreement rate between the LLM-based F1 metric $F_1^{\text{LM}}$ and human judgments, indicating strong alignment.

}

\section{Additional Experiments}

{
\subsection{Additional LLM-based Agentic Baselines}
\label{sec:appendix-additiona-llm-agent-baselines}

In addition to the simple 2-shot baseline, we profile two promising LLM-based agentic knowledge-base–construction baselines: HippoRAG~\citep{gutiérrez2024hipporag} and AutoSchemaKG~\citep{bai2025autoschemakgautonomousknowledgegraph}, representative of recent LLM-driven KG construction pipelines.

HippoRAG is a retrieval-augmented generation framework that builds a knowledge graph as an embedding index, mimicking the role of the hippocampus in human memory. We use the first stage of their KG construction pipeline, which consists of two prompts, one for named entity recognition (NER) and one for triple extraction. We also replace their 1-shot example with the same 2-shot examples used in our baseline.

AutoSchemaKG is a framework for web-scale KG construction over a pretraining-scale corpus. It calls three LLM modules on each webpage: (1) an entity–entity relationship extractor, (2) an entity–event relationship extractor, and (3) an event–event relationship extractor. These prompts are all zero-shot and are challenging to adapt, so we retain them as originally specified.

As shown in Table~\ref{tab:llm_metrics_all_baselines}, the simple 2-shot baseline outperforms both LLM-based baselines across all models evaluated on our task, including by more than 20\% for the strongest model, GPT-OSS-120B. Moreover, they inherit the same cost inefficiencies, as each webpage requires multiple LLM calls.
}

\subsection{Additional Ablation Experiment on Impact of Noisy Reward}
\label{sub:additional_ablation_noisy}

To further investigate the role of noisy reward, we conduct additional ablation experiments under three training configurations: (1) training on CC data only, (2) training on a mixture of CC and annotated data at a 1:1 ratio, and (3) training first on annotated data and then continuing on CC data. Results are reported in Table~\ref{tab:cc_ablation_noisy}.

\begin{table}[t]
\centering
\small
\resizebox{\textwidth}{!}{
\begin{tabular}{lccc|ccc|ccc}
\toprule
Method & \multicolumn{3}{c}{All}& \multicolumn{3}{c}{Example} & \multicolumn{3}{c}{Holdout} \\
 \cmidrule(lr){2-4}\cmidrule(lr){5-7}\cmidrule(lr){8-10}
 & $R^{\mathrm{LM}}$ & $P^{\mathrm{LM}}$ & $F_1^{\mathrm{LM}}$ & $R^{\mathrm{LM}}$ & $P^{\mathrm{LM}}$ & $F_1^{\mathrm{LM}}$ & $R^{\mathrm{LM}}$ & $P^{\mathrm{LM}}$ & $F_1^{\mathrm{LM}}$ \\
 \toprule
 Q-14B (Annotated mixed with CC) & 6.5 & 8.0 & 6.5 & 8.1 & 9.6 & 7.9 & 5.7 & 6.4 & 5.7 \\
 Q-14B (CC only) & 7.7 & 15.8  & 9.2  & 8.9 & 18.4  & 10.8 & 7.2 & 14.7 & 8.4 \\
 Q-14B (Annotated followed by CC) & \textbf{25.2} & \textbf{23.0} & \textbf{21.8} &  \textbf{34.9} & \textbf{31.0} & \textbf{30.0} & \textbf{20.5} & \textbf{19.1} & \textbf{17.7} \\
\bottomrule
\end{tabular}}
\caption{Ablation study on the impact of noisy reward. We compare three training configurations: (1) CC data only, (2) annotated data mixed with CC data at a 1:1 ratio, and (3) training first on annotated data followed by CC data. Results show that noisy reward alone or mixed training does not improve performance, whereas a staged setup, first training on annotated data before continuing with CC, yields substantial gains.}
\label{tab:cc_ablation_noisy}
\end{table}

{
\subsection{Additional Ablation Experiment on Domain Transferability}
\label{sec:appendix-ablation-domain}

In this ablation study, we reorganized the dataset by assigning each website to one of the following content categories:

\begin{itemize}
    \item Finance \& Economics
    \item Legal \& Regulatory
    \item Developer \& Software
    \item Science \& Research
    \item Science \& Database
    \item Sports
    \item Gaming \& Entertainment
    \item Media \& Entertainment
    \item Real Estate
    \item Social Platforms
    \item Weather \& Environment
    \item Jobs \& Careers
    \item Travel \& Hospitality
    \item Products \& Brands
    \item Encyclopedias \& Reference
\end{itemize}

We placed \textit{Products \& Brands} and \textit{Encyclopedias \& Reference} in the test set, with all remaining categories assigned to the training set. This split yielded 196 training examples and 72 test examples. After applying the maximum-context constraint described in Section~\ref{sec:appendix-data-proprocessing}, 147 training examples and 59 test examples remained.


}

\subsection{Complete Baseline Numbers}
\label{sub:appendiex_full_results}

For $F_1$, we provide two variants: (i) the macro-average of per-example $F_1$ scores, and (ii) a harmonic-mean variant defined as
\begin{equation}
\label{eq:harmonic_f1_def}
F_1^{H} = \frac{2\overline{P} \overline{R}}{\overline{P} + \overline{R}}
\end{equation}

where $\overline{P}$ and $\overline{R}$ denote the mean precision and recall, respectively. The complete list of baseline performance is shown in Table~\ref{tab:llm_metrics_all_baselines} and \ref{tab:llm_metrics_example_holdout_all}.


\begin{table}[t]
\centering
\small
\begin{tabular}{lcccc}
\toprule
Method & $R^{\mathrm{LM}}$ & $P^{\mathrm{LM}}$ & $F_1^{H, LM}$ & $F_1^{\mathrm{LM}}$ \\
\midrule
 \cellcolor{gray!10} & \cellcolor{gray!10} Baselines (Flattened) & \cellcolor{gray!10} & \cellcolor{gray!10} & \cellcolor{gray!10} \\
 {Q-14B w/ AutoSchemaKG~\citep{bai2025autoschemakgautonomousknowledgegraph}} & {2.1} & {8.26} & {3.35} & {8.17}\\
 {Q-14B w/ HippoRAG~\citep{gutiérrez2024hipporag} 2-shot} & {8.49} & {32.24} & {13.43} & {16.12}\\

 Q-14B flatten & 30.46 & 36.46 & 33.19 & 29.87 \\

{Q-32B w/ AutoSchemaKG~\citep{bai2025autoschemakgautonomousknowledgegraph}} & {2.64} & {11.14} & {4.27} & {9.33}\\
{Q-32B w/ HippoRAG~\citep{gutiérrez2024hipporag} 2-shot} & {10.12} & {39.7} & {16.13} & {20.03}\\
 Q-32B flatten & 28.73 & 37.44 & 32.51 & 29.93 \\
 {GO-20B w/ AutoSchemaKG~\citep{bai2025autoschemakgautonomousknowledgegraph}} & {5.57} & {11.96} & {7.6} & {9.26}\\
{GO-20B w/ HippoRAG~\citep{gutiérrez2024hipporag} 2-shot} & {8.26} & {23.06} & {12.16} & {14.02}\\
 GO-20B flatten & 36.94 & 37.88 & 37.40 & 33.61 \\
 GO-20B 2-shot flatten & 33.18 & 47.10 & 38.93 & 34.93 \\
{GO-120B w/ AutoSchemaKG~\citep{bai2025autoschemakgautonomousknowledgegraph}} & {6.52} & {16.97} & {9.42} & {12.28}\\
 {GO-120B w/ HippoRAG~\citep{gutiérrez2024hipporag} 2-shot} & {28.57} & {12.12} & {17.02} & {17.22}\\
 GO-120B flatten & 36.43 & 34.59 & 35.49 & 31.74 \\
 GO-120B 2-shot flatten & 42.27 & 46.26 & 44.18 & 40.40 \\

\midrule
 \cellcolor{gray!10} & \cellcolor{gray!10} Baselines (Script-gen) & \cellcolor{gray!10} & \cellcolor{gray!10} & \cellcolor{gray!10} \\
Q-14B  agentic-3-iter & 8.11 & 8.26 & 8.18 & 7.14 \\
Q-14B  agentic-3-iter 2-shot & 8.59 & 11.13 & 9.70 & 8.01 \\
Q-32B  agentic-3-iter & 10.41 & 9.08 & 9.70 & 8.74 \\
Q-32B  agentic-3-iter 2-shot & 18.56 & 27.20 & 22.07 & 19.41 \\
Q-72B  agentic-3-iter & 9.67 & 9.65 & 9.66 & 7.19 \\
Q-72B  agentic-3-iter 2-shot & 16.40 & 19.41 & 17.78 & 14.97 \\
GO-20B  agentic-3-iter & 24.70 & 23.22 & 23.94 & 20.87 \\
GO-20B  agentic-3-iter 2-shot & 13.06 & 27.30 & 17.66 & 14.40 \\
GO-120B  agentic-3-iter & 27.63 & 24.76 & 26.12 & 23.30 \\
GO-120B  agentic-3-iter 2-shot & 33.86 & 40.96 & 37.07 & 34.30 \\
GPT-4o  agentic-3-iter & 19.05 & 14.72 & 16.61 & 13.81 \\
GPT-4o  agentic-3-iter 2-shot & 25.95 & 33.04 & 29.07 & 24.42 \\
L-70B  agentic-3-iter & 10.05 & 15.49 & 12.19 & 10.47 \\
L-70B  agentic-3-iter 2-shot & 8.23 & 8.08 & 8.15 & 7.10 \\

\midrule
\cellcolor{blue!10} & \cellcolor{blue!10} \system & \cellcolor{blue!10} & \cellcolor{blue!10} & \cellcolor{blue!10} \\


Q-14B & 22.96 & 24.26 & 23.59 & 19.91 \\
Q-14B (+CC) & 25.24 & 22.98 & 24.05 & 21.77 \\

Q-32B & 29.88 & 31.53 & 30.68 & 28.05 \\
Q-32B (+CC) & 37.41 & 36.03 & 36.71 & 33.24 \\


\bottomrule


\end{tabular}
\caption{List of all baselines and \system-trained models. LLM-judged metrics on all data. $P^{\mathrm{LM}}$, $R^{\mathrm{LM}}$, harmonic $F_1^{H, LM}$, and average per-example $F_1^{\mathrm{LM}}$.}
\label{tab:llm_metrics_all_baselines}
\end{table}

\begin{table}[t]
\centering
\small
\resizebox{\textwidth}{!}{
\begin{tabular}{lcccccccc}
\toprule
Method & \multicolumn{4}{c}{Example} & \multicolumn{4}{c}{Holdout} \\
 \cmidrule(lr){2-5}\cmidrule(lr){6-9}
 & $R^{\mathrm{LM}}$ & $P^{\mathrm{LM}}$ & $F_1^{H, LM}$ & $F_1^{\mathrm{LM}}$ & $R^{\mathrm{LM}}$ & $P^{\mathrm{LM}}$ & $F_1^{H, LM}$ & $F_1^{\mathrm{LM}}$ \\
\midrule
 \cellcolor{gray!10} & \cellcolor{gray!10} & \cellcolor{gray!10} & \cellcolor{gray!10} Baselines  & \cellcolor{gray!10} & \cellcolor{gray!10} & \cellcolor{gray!10} & \cellcolor{gray!10} & \cellcolor{gray!10} \\
Q-14B  agentic-3-iter & 11.96 & 11.81 & 11.88 & 10.57 & 6.47 & 6.90 & 6.68 & 5.77 \\
Q-14B  agentic-3-iter 2-shot & 13.21 & 17.97 & 15.23 & 12.63 & 6.29 & 7.79 & 6.96 & 5.73 \\
Q-32B  agentic-3-iter & 18.84 & 17.17 & 17.97 & 16.46 & 6.36 & 5.33 & 5.80 & 5.07 \\
Q-32B  agentic-3-iter 2-shot & 24.53 & 34.83 & 28.79 & 25.90 & 15.79 & 23.91 & 19.02 & 16.40 \\
Q-72B  agentic-3-iter & 13.03 & 13.15 & 13.09 & 10.12 & 8.20 & 8.12 & 8.16 & 5.94 \\
Q-72B  agentic-3-iter 2-shot & 24.11 & 28.59 & 26.16 & 21.78 & 13.26 & 15.83 & 14.43 & 12.38 \\
GO-20B  agentic-3-iter & 29.25 & 26.38 & 27.74 & 24.91 & 22.51 & 21.78 & 22.14 & 18.94 \\
GO-20B  agentic-3-iter 2-shot & 13.48 & 27.68 & 18.13 & 14.41 & 13.07 & 27.11 & 17.64 & 14.66 \\
GO-120B  agentic-3-iter & 31.32 & 26.76 & 28.86 & 25.70 & 25.86 & 23.86 & 24.82 & 22.16 \\
GO-120B  agentic-3-iter 2-shot & 35.83 & 42.27 & 38.78 & 36.60 & 32.98 & 40.47 & 36.34 & 33.26 \\
GPT-4o  agentic-3-iter & 25.19 & 18.35 & 21.23 & 18.47 & 16.00 & 12.89 & 14.28 & 11.47 \\
GPT-4o  agentic-3-iter 2-shot & 32.98 & 36.48 & 34.64 & 31.19 & 22.52 & 31.32 & 26.20 & 21.11 \\
L-70B  agentic-3-iter & 16.65 & 23.76 & 19.58 & 16.78 & 6.86 & 11.16 & 8.49 & 7.36 \\
L-70B  agentic-3-iter 2-shot & 7.77 & 6.77 & 7.23 & 6.18 & 8.42 & 8.68 & 8.54 & 7.51 \\

\midrule
\cellcolor{blue!10} & \cellcolor{blue!10}  & \cellcolor{blue!10} & \cellcolor{blue!10}\system & \cellcolor{blue!10} & \cellcolor{blue!10} & \cellcolor{blue!10} & \cellcolor{blue!10} & \cellcolor{blue!10}\\

Q-14B & 31.22 & 29.81 & 30.50 & 26.71 & 19.01 & 21.65 & 20.24 & 16.66 \\
Q-14B (+CC) & 34.88 & 30.96 & 32.80 & 29.96 & 20.45 & 19.06 & 19.73 & 17.69 \\


Q-32B & 31.99 & 33.88 & 32.90 & 30.32 & 28.79 & 30.28 & 29.51 & 26.83 \\
Q-32B (+CC) & 39.54 & 35.48 & 37.40 & 34.60 & 36.24 & 36.15 & 36.20 & 32.41 \\


\bottomrule
\end{tabular}}
\caption{List of all baselines and \system-trained models by Example and Holdout. LLM-judged metrics on all data. $P^{\mathrm{LM}}$, $R^{\mathrm{LM}}$, harmonic $F_1^{H, LM}$, and average per-example $F_1^{\mathrm{LM}}$.}
\label{tab:llm_metrics_example_holdout_all}
\end{table}

{
\section{Details on QA dataset used in Sec. \ref{sec:qa-improvement}}

The QA pairs in Sec. \ref{sec:qa-improvement} are collected by \citet{kai-2025-arxiv} through an LLM-generated followed by human-auditing process. We summarize their process below:

First, a 70B Llama model generated initial question--answer pairs using webpage content and ground truth data. Then, these pairs were refined by:
\begin{enumerate}
    \item \textbf{Removing overly complex questions} that required heavy reasoning, focusing instead on comprehension of semi-structured webpages.
    \item \textbf{Eliminating compound questions} that combined multiple queries into one to avoid inflated difficulty.
    \item \textbf{Filtering out trivial questions} that all tested models answered correctly, ensuring better differentiation among model performances.
\end{enumerate}

Finally, \textbf{human auditors} reviewed and removed any pairs that were ungrounded in the source content or contained incorrect answers.

For more details and statistics, refer to \citet{kai-2025-arxiv}.
}

{
\section{Example Model Predictions}
\label{sec:appendix-example-predictions}

Here, we show an example of our best model's prediction on the eval set to better illustrate the task and associated difficulties.

For \href{https://web.archive.org/web/20250104135850/https://www.sec.gov/Archives/edgar/data/1000069/000089418910002875/0000894189-10-002875-index.htm}{\textcolor{blue}{\underline{this website}}}, our model generated the code shown in Table~\ref{code:example-1}. The prediction versus gold entries are shown in Table~\ref{tab:triple_comparison-1}. This code achieves the following scores: $R^{\text{LM}} = 53.3,  P^{\text{LM}} =50.0, F_1^{\text{LM}}=51.6$.

The model code misses some information such as the ``Business Address'' and ``Mailing Address'' fields, penalizing the recalls. It also produces some false positives, including triples where the relation is “sequence” but no corresponding object is returned, reducing precision. Additionally, the model incorrectly treats the filing document number as the subject throughout, rather than using more appropriate subjects such as ``Class/Contract C000008114.''

}

\begin{table*}
\begin{lstlisting}[style=python]
from bs4 import BeautifulSoup
from typing import List

def main(html: str) -> List[List[str]]:
    soup = BeautifulSoup(html, 'html.parser')
    triples = []
    
    # Extract title
    title = soup.title.text if soup.title else ""
    triples.append(["EDGAR Filing", "title", title])
    
    # Extract form information
    form_name = soup.find("div", id="formName")
    if form_name:
        triples.append([title, "form", form_name.text.strip()])
    
    sec_num = soup.find("div", id="secNum")
    if sec_num:
        triples.append([title, "accessionNumber", sec_num.text.strip()])
    
    # Extract filing details
    info_heads = soup.find_all("div", class_="infoHead")
    info_values = soup.find_all("div", class_="info")
    for head, value in zip(info_heads, info_values):
        triples.append([title, head.text.strip(), value.text.strip()])
    
    # Extract document table information
    table = soup.find("table", class_="tableFile")
    if table:
        rows = table.find_all("tr")
        for row in rows[1:]:  # Skip header row
            cols = row.find_all(["td", "th"])
            if len(cols) >= 5:
                seq = cols[0].text.strip()
                description = cols[1].text.strip()
                document = cols[2].text.strip()
                doc_type = cols[3].text.strip()
                size = cols[4].text.strip()
                triples.append([title, "sequence", seq])
                triples.append([title, "description", description])
                triples.append([title, "document", document])
                triples.append([title, "type", doc_type])
                triples.append([title, "size", size])
    
    # Extract series and classes information
    series_table = soup.find("table", class_="tableSeries")
    if series_table:
        rows = series_table.find_all("tr")
        for row in rows[2:]:  # Skip header rows
            cols = row.find_all(["td", "th"])
            if len(cols) >= 4:
                status = cols[0].text.strip()
                name = cols[2].text.strip()
                ticker = cols[3].text.strip()
                triples.append([title, "status", status])
                triples.append([title, "name", name])
                triples.append([title, "tickerSymbol", ticker])
    
    return triples
\end{lstlisting}
\caption{{An example SCRIBES-generated code discussed in Appendix~\ref{sec:appendix-example-predictions}.}}
\label{code:example-1}
\end{table*}

\begin{table}[t]
\centering
{ \scriptsize
\resizebox{\textwidth}{!}{%
\begin{tabular}{p{0.42\linewidth} p{0.16\linewidth} p{0.42\linewidth}}
\hline
Subject & Relation & Object \\
\hline
\multicolumn{3}{l}{\textbf{Gold}} \\
 EDGAR Filing Documents for 0000894189-10-002875 & Filing Date & 2010-08-10 \\
 EDGAR Filing Documents for 0000894189-10-002875 & Accepted & 2010-08-10 10:58:25 \\
 EDGAR Filing Documents for 0000894189-10-002875 & Documents & 2 \\
 EDGAR Filing Documents for 0000894189-10-002875 & Period of Report & 2010-06-30 \\
 EDGAR Filing Documents for 0000894189-10-002875 & Effectiveness Date & 2010-08-10 \\
 empiric\_63010nq.htm & Seq & 1 \\
 empiric\_63010nq.htm & Description & QUARTERLY NOTICE OF PORTFOLIO HOLDINGS \\
 empiric\_63010nq.htm & Type & N-Q \\
 empiric\_63010nq.htm & Size & 530257 \\
 certs.htm & Seq & 2 \\
 certs.htm & Description & OFFICER CERTIFICATIONS \\
 certs.htm & Type & EX-99.CERT \\
 certs.htm & Size & 24222 \\
 0000894189-10-002875.txt & Description & Complete submission text file \\
 0000894189-10-002875.txt & Size & 556496 \\
 Series S000002964 & Name & Core Equity Fund \\
 Class/Contract C000008114 & Name & C \\
 Class/Contract C000008114 & Ticker Symbol & EMCCX \\
 Class/Contract C000008115 & Name & A \\
 Class/Contract C000008115 & Ticker Symbol & EMCAX \\
 EDGAR Filing Documents for 0000894189-10-002875 & EMPIRIC FUNDS, INC (Filer) CIK & 0001000069 (see all company filings) \\
 EDGAR Filing Documents for 0000894189-10-002875 & IRS No. & 742759654 \\
 EDGAR Filing Documents for 0000894189-10-002875 & State of Incorp. & TX \\
 EDGAR Filing Documents for 0000894189-10-002875 & Fiscal Year End & 931 \\
 EDGAR Filing Documents for 0000894189-10-002875 & Type & N-Q \\
 EDGAR Filing Documents for 0000894189-10-002875 & Act & 40 \\
 EDGAR Filing Documents for 0000894189-10-002875 & File No. & 811-09088 \\
 EDGAR Filing Documents for 0000894189-10-002875 & Film No. & 101003905 \\
 EDGAR Filing Documents for 0000894189-10-002875 & Business Address & 6300 BRIDGEPOINT PARKWAY BUILDING II, SUITE 105 AUSTIN TX 78730 5123289321X1 \\
 EDGAR Filing Documents for 0000894189-10-002875 & Mailing Address & 6300 BRIDGEPOINT PARKWAY BUILDING II, SUITE 105 AUSTIN TX 78730 \\
\hline
\multicolumn{3}{l}{\textbf{Predicted}} \\
 EDGAR Filing & title & EDGAR Filing Documents for 0000894189-10-002875 \\
 EDGAR Filing Documents for 0000894189-10-002875 & form & Form N-Q - Quarterly Schedule of portfolio holdings of management investment companies: \\
 EDGAR Filing Documents for 0000894189-10-002875 & accessionNumber & SEC Accession No. 0000894189-10-002875 \\
 EDGAR Filing Documents for 0000894189-10-002875 & Filing Date & 2010-08-10 \\
 EDGAR Filing Documents for 0000894189-10-002875 & Accepted & 2010-08-10 10:58:25 \\
 EDGAR Filing Documents for 0000894189-10-002875 & Documents & 2 \\
 EDGAR Filing Documents for 0000894189-10-002875 & Period of Report & 2010-06-30 \\
 EDGAR Filing Documents for 0000894189-10-002875 & Effectiveness Date & 2010-08-10 \\
 EDGAR Filing Documents for 0000894189-10-002875 & sequence & 1 \\
 EDGAR Filing Documents for 0000894189-10-002875 & description & QUARTERLY NOTICE OF PORTFOLIO HOLDINGS \\
 EDGAR Filing Documents for 0000894189-10-002875 & document & empiric\_63010nq.htm \\
 EDGAR Filing Documents for 0000894189-10-002875 & type & N-Q \\
 EDGAR Filing Documents for 0000894189-10-002875 & size & 530257 \\
 EDGAR Filing Documents for 0000894189-10-002875 & sequence & 2 \\
 EDGAR Filing Documents for 0000894189-10-002875 & description & OFFICER CERTIFICATIONS \\
 EDGAR Filing Documents for 0000894189-10-002875 & document & certs.htm \\
 EDGAR Filing Documents for 0000894189-10-002875 & type & EX-99.CERT \\
 EDGAR Filing Documents for 0000894189-10-002875 & size & 24222 \\
 EDGAR Filing Documents for 0000894189-10-002875 & sequence &  \\
 EDGAR Filing Documents for 0000894189-10-002875 & description & Complete submission text file \\
 EDGAR Filing Documents for 0000894189-10-002875 & document & 0000894189-10-002875.txt \\
 EDGAR Filing Documents for 0000894189-10-002875 & type &  \\
 EDGAR Filing Documents for 0000894189-10-002875 & size & 556496 \\
 EDGAR Filing Documents for 0000894189-10-002875 & status & CIK 0001000069 \\
 EDGAR Filing Documents for 0000894189-10-002875 & name &  \\
 EDGAR Filing Documents for 0000894189-10-002875 & tickerSymbol &  \\
 EDGAR Filing Documents for 0000894189-10-002875 & status & Class/Contract C000008114 \\
 EDGAR Filing Documents for 0000894189-10-002875 & name & C \\
 EDGAR Filing Documents for 0000894189-10-002875 & tickerSymbol & EMCCX \\
 EDGAR Filing Documents for 0000894189-10-002875 & status & Class/Contract C000008115 \\
 EDGAR Filing Documents for 0000894189-10-002875 & name & A \\
 EDGAR Filing Documents for 0000894189-10-002875 & tickerSymbol & EMCAX \\
\hline
\end{tabular}
}
}
\caption{{Comparison of predicted and gold triples for Code \ref{code:example-1}.}}
\label{tab:triple_comparison-1}
\end{table}

\section{Prompts Used}

All prompts used in our experiments are shown here in Jinja2 format, including the classifier prompt (Prompt \ref{prompt:classifier}), LLM direct extraction prompt (Prompt \ref{prompt:direct-gen}), LLM-as-a-judge prompt (Prompt \ref{prompt:llm-as-a-judge}), QA prompt (Prompt \ref{prompt:qa-reference}), the main script generation prompt (Prompt \ref{prompt:script-gen}) used in both baseline and in \system training data, and the QA evaluation prompt (Prompt \ref{prompt:qa-eval}).

\begin{table*}
\begin{lstlisting}[basicstyle=\ttfamily\tiny]
# instruction

Your task is to classify an input HTML to see whether it contains semi-structured content.

You are shown below with one example with semi-structured content and one without.
Output a JSON with the following two fields: "reason" and "decision".
Reason should specify your chain of thought and decision should be one of:

- Semi-structured content: Respond with "Yes" if the HTML contains semi-structured content,
such as tables and infoboxes.
- No semi-structured content: Respond with "No" if the HTML does not contain any semi-structured content.
- Explicit content: Respond with "Exclude" if the HTML contains explicit content
(e.g., adult material, graphic violence).

# input

Exaples containing the following HTML:

{{ HTML_example_1 }}

# output

{
    "reason": "This HTML contains a table which falls into the definition of semi-structured content",
    "decision": "Yes"
}

# input

{{ HTML_example_2 }}

# output

{
    "reason": "Even though this HTML contains structured discussions and Q&As, it does not have tables or infoboxes",
    "decision": "No"
}


# input

An HTML with the following info:

{{ HTML_example_3 }}

# output


{
    "reason": "This HTML show cases a infobox, which should be treated as a semi-structured content.",
    "decision": "Yes"
}

# input

{{ html }}

\end{lstlisting}
\caption{Classifier prompt used to determine whether a webpage contains semi-structured content or not.}
\label{prompt:classifier}
\end{table*}

\begin{table*}
\begin{lstlisting}[basicstyle=\ttfamily\tiny]
# instruction
You are given a doc in HTML and its title. Please return all (subject, predicate, object) triples
that can be extracted from the doc, in the order they appear in the doc. For large chunk of descriptions
or sections of free-form text, you should keep them as object. Do not attempt to break big chunks
of texts down into smaller portions.

Subject, predicate, and object should generally be gained from the text spans in the doc or the title.
Please only include complete triples; if for any section the predicate or object is missing from the doc,
you may skip it.
Output a list of lists, where each inner list is a triple. I will use python's eval to parse your output.

# input
{% if example_global_html_triples %}
Here are {{ example_global_html_triples|length }} examples of flattened HTML pages and their expected triples:
{% for single_example in example_global_html_triples %}
Example {{ loop.index0 }} Flattened HTML: {{ single_example["html_flatten"] }}
Example {{ loop.index0 }} Expected Triples: {{ single_example["triples_annotation"] }}
{% endfor %}
{% endif %}

{% if example_triples %}
Here are 10 triples we are expecting in the output randomly chosen: {{ example_triples }}
{% endif %}

### title
{{ html_title }}

### HTML
{{ html }}

\end{lstlisting}
\caption{LLM direct extraction prompt used to directly generate triples from a webpage.}
\label{prompt:direct-gen}
\end{table*}

\begin{table*}
\begin{lstlisting}[basicstyle=\ttfamily\tiny]
# instruction

You are given two (subject, predicate, object) triples.
Your response should be "Yes" if the triples are semantically the same or "No"
if they are semantically different.

# input
{{ tx }}
{{ ty }}


\end{lstlisting}
\caption{LLM-as-a-judge prompt for judging whether two triples are semantically equivalent.}
\label{prompt:llm-as-a-judge}
\end{table*}

\begin{table*}
\begin{lstlisting}[basicstyle=\ttfamily\tiny]
# instruction
You are given a question and a reference that may or may not help answer the question.
Please answer the question. Be concise.

# input
### Question
{{ question }}
### Reference
{{ reference }}



\end{lstlisting}
\caption{Question Answering prompt with reference.}
\label{prompt:qa-reference}
\end{table*}

\begin{table*}
\begin{lstlisting}[basicstyle=\ttfamily\tiny]
# instruction

Your task is to generate semantic triples from a given HTML.
A triple contains a subject, a predicate, and an object.
You should write python code to extract triples from the HTML.
The final executable function should be called `def main(html) -> List[tuple(str, str, str)]:`,
where it will output a list of triples.
You should output the python code only. Feel free to add comments to explain your code.
Do not include any text other than the code in your response.

IMPORTANT: we will re-use the same script for other webpages with similar HTML contents.
So you should make your script re-usable across different websites
(do not hardcode for values for this particular HTML).

# input

{% if example_global_html_triples %}
Here are {{ example_global_html_triples|length }} examples of other HTML sites and
what the script-generated output we are looking for:
{% for single_example in example_global_html_triples %}
Example {{ loop.index0 }} HTML: {{ single_example["html_content"]  }}
Example {{ loop.index0 }} Expected Outputs: {{ single_example["triples_annotation"] }}
{% endfor %}
{% endif %}

HTML: {{ html }}

{% if example_triples %}
Here are 10 triples we are expecting in the output randomly chosen: {{ example_triples }}
{% endif %}
{% if all_triples %}
Here are all the triples we are expecting in the output: {{ all_triples }}
{% endif %}

{% if prev_script %}
You previously generated a script:
{{ prev_script }}

This script generated the following result:
{{ feedback }}

If you think the results are good enough, stop and output the same script.
If not, incorporate the feedback in generating a new script.
{% endif %}


\end{lstlisting}
\caption{Main script generation prompt for baselines and \system-trained models.}
\label{prompt:script-gen}
\end{table*}

\begin{table*}
\begin{lstlisting}[basicstyle=\ttfamily\tiny]

# instruction

You need to check whether the prediction of a question-answering system to a question is correct.
You should make the judgment based on the ground truth answer provided to you.

Your response should be "correct" if the prediction is correct or "incorrect" if the prediction is wrong.

# input

Question: {{ question }}
Ground truth: {{ gold }}
Prediction: {{ answer }}
Correctness:

\end{lstlisting}
\caption{QA evaluation prompt.}
\label{prompt:qa-eval}
\end{table*}

\end{document}